\documentclass{article}

     \usepackage[nonatbib,preprint]{neurips_2020}

\usepackage[utf8]{inputenc} % allow utf-8 input
\usepackage[T1]{fontenc}    % use 8-bit T1 fonts
\usepackage{hyperref}       % hyperlinks
\usepackage{url}            % simple URL typesetting
\usepackage{booktabs}       % professional-quality tables
\usepackage{amsfonts}       % blackboard math symbols
\usepackage{nicefrac}       % compact symbols for 1/2, etc.
\usepackage{microtype}      % microtypography
\usepackage{subcaption}
\usepackage{multirow}
\usepackage{multicol}
\usepackage{amsbsy}
\usepackage[shortlabels]{enumitem}

\usepackage{todonotes}

\usepackage{rotating}

\title{A Comprehensive Evaluation of Multi-task Learning and Multi-task Pre-training on EHR Time-series Data}

\author{
Matthew B. A. McDermott\thanks{Equal Contribution.} \\
CSAIL, MIT \\
\texttt{mmd@mit.edu}\\
\And
Bret Nestor$^*$ \\
University of Toronto \\
\And 
Evan Kim \\
CSAIL, MIT \\
\And 
Wancong Zhang \\
NYU \\
\And 
Anna Goldenberg \\
University of Toronto, Vector Institute, SickKids \\
\And 
Peter Szolovits \\
CSAIL, MIT \\ 
\And
Marzyeh Ghassemi \\
University of Toronto, Vector Institute
}

\begin{document}
% \todo{check citation format}
% \todo{move figure captions to lower right in Figure 4}

\maketitle

\begin{abstract}
Multi-task learning (MTL) is a machine learning technique aiming to improve model performance by leveraging information across many tasks. It has been used extensively on various data modalities, including electronic health record (EHR) data. However, despite significant use on EHR data, there has been little systematic investigation of the utility of MTL across the diverse set of possible tasks and training schemes of interest in healthcare. In this work, we examine MTL across a battery of tasks on EHR time-series data. We find that while MTL does suffer from common negative transfer, we can realize significant gains via MTL pre-training combined with single-task fine-tuning. We demonstrate that these gains can be achieved in a task-independent manner and offer not only minor improvements under traditional learning, but also notable gains in a few-shot learning context, thereby suggesting this could be a scalable vehicle to offer improved performance in important healthcare contexts.
\end{abstract}

\section{Introduction}
Multi-task learning (MTL) is a machine learning technique aiming to improve model performance by leveraging information across many tasks. MTL has been explored extensively, especially in the computer vision and natural language processing domains~\cite{Caruana_mtl,luong2015multi,zhang2017survey}. Research has found that MTL can offer performance benefits for similar tasks, while for dissimilar tasks, it may induce \emph{negative transfer}, where MTL harms overall performance~\cite{ndirango2019generalization, Wu2020Understanding}. Additionally, some argue that MTL can act as a regularizer in learning~\cite{ruder2017overview}. Others have noted that MTL can induce gains in  pre-training/single-task fine-tuning, few-shot learning, or fairness contexts~\cite{DBLP:conf/interspeech/HeLYHC18, das2018mitigating, liu2019multi, tian2020rethinking, mtlearning_fair_2019}. 

One domain where MTL may be particularly helpful is machine learning for health (ML4H), a field which suffers from notable data difficulties that motivate the use of MTL.
Clinical data is often smaller, higher-dimensional, and noisier than general domain data, and tasks are commonly susceptible to confounders that vary across institutions, time, and demographics~\cite{pmlr-v106-prabhu19a,ghassemi2019practical,wiens2019no, counterfactual_norm,pmlr-v106-nestor19a,PMID:29893864}, which may be alleviated by MTL's regularizing effect.
Additionally, clinical data poses novel challenges, such as the prevalence of diverse rare diseases~\cite{2019arXiv191113232C,wakap2020estimating, 10.1001/jamanetworkopen.2020.1965}, for which data is scarce and MTL's possible benefits for few-shot learning would be critical,
or the importance of model fairness~\cite{10.1001/jamainternmed.2018.3763,Obermeyer447},
where MTL's possible ability to yield more equitable models under imbalanced data would be very valuable. 
These data difficulties impose substantial hurdles to effective learning in this domain and motivate the use and understanding of MTL in this context.
While MTL has been used in ML4H~\cite{harutyunyan_multitask_2017, suresh_learning_2018, Si2019}, a general understanding of its broader efficacy on clinical time-series is lacking. Further, important use cases of MTL, such as its ability to aid diverse tasks in a pre-training/fine-tuning setup, its efficacy for few-shot learning, and its advantages on imbalanced datasets have yet to be studied.

In this work, we provide a robust analysis of MTL across various learning contexts in ML4H. We design a broad set of tasks over physiological electronic health record (EHR) time-series data and use them to answer the following specific questions:
% First, how extensive is negative transfer among commonly studied ML4H tasks? Second, can MTL offer any general performance benefits when used either directly or with pre-trained models? Following that, do these benefits extend to few-shot learning contexts, when we wish to generalize beyond a pre-trained MTL representation to a very small fine-tuning dataset? And, lastly, can MTL pre-training help reduce bias and improve fairness in downstream tasks?
\begin{enumerate}[nosep]
    \item How extensive is negative transfer among traditionally studied tasks?
    \item Can MTL consistently offer benefits in performance?
    \item Can MTL pre-training offer benefits for few-shot learning?
    \item Can MTL pre-training help reduce sensitivity to population imbalance?
\end{enumerate}

Our analyses reveal that negative transfer is common, and conventional MTL often does not yield performance benefits. In contrast, we observe that MTL pre-training followed by single-task fine-tuning does match or exceed both single-task and multi-task training, and offers significant gains in the few-shot regime. Unfortunately, we do not find evidence that it remedies issues of population imbalance on the chosen demographics. Overall, these findings robustly suggest that multi-task pre-training, followed by task-specific fine-tuning, can offer advantages in machine learning for health, particularly in few-shot or small data contexts.

\section{Related Works}
% \todo{add
% % mihaela
% sontag?
% % sun
% % succhi
% finale,
% % succhi lung trajectory citations
% }
% could also mention  jimeng
% \todo{sepsis \cite{futoma2017improved, Henry299ra122}}
Prior work has demonstrated the benefits of MTL in various contexts \cite{10.1023/A:1007379606734,10.1145/1014052.1014067,NIPS2006_3143}, including in deep neural models specifically for sequence learning \cite{luong2015multitask}, face detection \cite{8170321}, and self-supervised learning \cite{Doersch_2017_ICCV}. MTL has also been explored specifically in EHR data, including using MTL to improve mortality, readmission, long length-of-stay, or phenotype (ICD code) prediction~\cite{harutyunyan_multitask_2017}, in-hospital, 30-day, and 1-year mortality prediction directly~\cite{Si2019}, molecular property prediction~\cite{wu2018moleculenet}, or adverse outcome prediction across different patient sub-populations~\cite{suresh_learning_2018}. MTL has also been used address incomplete health data~\cite{hunt2018multi} or rare disease detection~\cite{liu2020multi,2019arXiv191113232C}, and has been generally applied to many specific ML4H questions~\cite{soleimani2017scalable,zhou2013modeling,schulam2015framework,alaa2017bayesian,futoma2017improved}. 

Despite the widespread use of MTL in ML4H, careful analyses of its use outside direct performance comparisons on the tasks under study has been limited. Ding et al.~\cite{ding2019effectiveness} offer an assessment on the propensity of negative transfer within the context of ICD code prediction, finding that negative transfer can occur on a subset of tasks even while more rare tasks still obtain a benefit, but similar analyses beyond ICD-code prediction tasks do not exist.

Transfer learning has been used extensively in the medical imaging domain; however, only recently have researchers performed focused investigation on the efficacy of transfer learning over medical imaging in general~\cite{raghu2019transfusion}.
% however, recent evidence has suggested that for overparamaterised models, transfer learning from the general domain to medical imaging models may recycle features leading to improved performance and rapid convergence despite domain differences~\cite{raghu2019transfusion}.
Pre-training has also been applied on EHR data, largely by using self-supervised autoregressive pre-training tasks~\cite{shang_pre-training_2019,li2020behrt}. In this work, we perform a more robust analysis of these techniques, moving to establish best practices.

\section{Methods}
\begin{figure}
    \centering
    \begin{subfigure}{0.275\linewidth}
        \centering
        \includegraphics[width=\linewidth]{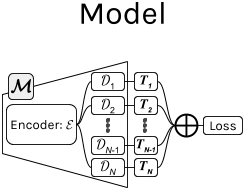}
        \caption{Our model $\mathcal M$ is composed of an encoder $\mathcal E$ which is used for all $N$ tasks, followed by task-specific decoders $\mathcal D_1,\ldots, \mathcal D_N$. In certain configurations (e.g., single-task training) only a subset of tasks are used in training.}
        \label{subfig:model}
    \end{subfigure}
    \hfill
    \begin{subfigure}{0.675\linewidth}
        \centering
        \includegraphics[width=\linewidth]{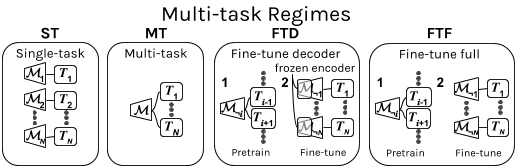}
        \caption{We use several multi-task regimes in this study, including single-task (\textbf{ST}), in which a separate model $\mathcal M_t$ is trained on each task $t$, multi-task (\textbf{MT}), in which a single model $\mathcal M$ is trained on all tasks simultaneously, and two fine-tuning models, fine-tune decoder (\textbf{FTD}) and fine-tune full (\textbf{FTF}), both of which start with \emph{task-omitted} pre-training, where a model $\mathcal M_t$ is trained on all tasks \emph{except} for $t$, followed by fine-tuning of $\mathcal M_t$ on task $t$ alone, either allowing only the decoder $\mathcal D_t$ to specialize in finetuning (FTD) or allowing both the encoder $\mathcal E$ and the decoder $\mathcal D_t$ to specialize (FTF).
        }
        \label{subfig:MTL_regimes}
    \end{subfigure}
    
    \caption{Our model and multi-task regimes.}
    \label{fig:model_overview}
    
    % \includegraphics[width=1.0\linewidth]{Images/overview.png}
    
    % \caption{
    % We train a multi-task model $\mathcal M$ comprises a shared encoder, and several task specific decoders $\mathcal D_i$, across a set of tasks with labels $T_1, \ldots, T_N$, in one of several regimes enabling us to probe the efficacy of MTL. 
    % % An overview of our model architecture. Numerical signals, including labs, vitals, and continuous encodings of static, demographic data, along with treatments, are imputed, then ingested into a single, multi-task encoder to yield a shared representation. Next, this representation is decoded by task specific decoders into each of our tasks, whose losses are aggregated during training.\todo{Define symbols$\mathcal{M, D, T}$}
    % }
    % \label{fig:model_overview}
\end{figure}

\subsection{Model Composition}
We use a shared encoder sub-network, which is shared across all tasks, and separate, independent per-task decoder heads for actual prediction. More formally, given a sample from a dataset of EHR physiological timeseries $\vec x_i$ and $N$ tasks with labels given by $T_1^{(i)}, \ldots, T_N^{(i)}$, we train a multi-task model $\mathcal M$ structured via a shared encoder $\mathcal E$ followed by $N$ task-specific decoder modules $\mathcal D_1, \ldots, \mathcal D_N$, such that the prediction of model $\mathcal M$ on task $t$ is given by $\mathcal D_t(\mathcal E(\vec x_i))$. Individual losses are computed for each task, which are then summed across all tasks to form the overall learning loss: $\mathcal L(\vec x_i, \vec {T}^{(i)}) = \sum_{j=1}^N \mathcal L_j(T^{(i)}_j, \mathcal D_j(\mathcal E(\vec x_i)))$, and learning over both encoder parameters and decoder parameters is performed via the Adam~\cite{DBLP:journals/corr/KingmaB14} variant of stochastic gradient descent over $\mathcal L$. This setup is shown in Figure~\ref{subfig:model}.

\subsection{Multi-task Regimes}
\label{sec:training_regimes}

In this work, recall that we are interested in answering 4 key questions about the efficacy of MTL over clinical data: how extensive is negative transfer, does MTL offer any direct performance benefits, and can MTL pre-training help us improve few-shot learning or ameliorate bias. To assess each of these, we need to use several MTL regimes, which we detail here and show graphically in Figure~\ref{subfig:MTL_regimes}.

\subsubsection{Single-task (ST)}
As a baseline we train an unconstrained (encoder \& decoder) model for each task independently, optimizing $\mathcal L_t$ individually for each task $t$. Note that for single-task training, we retain the same structure of our model, but as we have a separate model $\mathcal M_t$ for each task $t$, the encoder and decoder are both functionally task-specific. These experiments will serve as our comparison point for a global assessment of negative transfer, in which we compare a full MTL system trained across all tasks to the ST results independently, as well as our comparison point for the MTL pre-training system across raw, few-shot, and biased data.

\subsubsection{Full Multi-task (MT)}
We train unconstrained (encoder \& decoder) models in a conventional multi-task setting across all tasks. These results are used as comparison points for single-task and fine-tuned experiments.

\subsubsection{Pre-training/Fine-tuning}
\label{subsec:pre-training}

\paragraph{Task-Omitted Pre-training}
Prior to fine-tuning, we must pretrain a model with the task-of-interest omitted.\footnote{We don't pre-train with the task of interest included as we want to simulate fine-tuning on a completely unseen task, to assess the utility of MTL pre-training in ML4H for use across diverse (potentially unknown) downstream tasks, rather than on a static collection of tasks of interest.} For each task $t$, we train a model $\mathcal M_{\neg t}$ in a multi-task setting on all tasks \emph{except} for task $t$ by optimizing over the loss $\mathcal L_{\neg t} = \sum_{j \neq t} \mathcal L_j$. These models are then saved and used for several purposes. First, these models' performance on all tasks still included in the training ensemble (e.g., all tasks other than $t$) can offer us a \emph{local} picture of the extent of negative transfer, by allowing us to compare to the full MT results and judge how removing a single task affects the performance on the other tasks in the ensemble. This complements our more global comparison to the single-task system, which dictates how the inclusion of all other tasks affects performance. Second, these models are used as our pre-trained sources for our analyses into the effect of MTL pre-training on direct performance, few-shot learning, and imbalanced data, so that we can simulate adapting a multi-task model to a completely new task.

\paragraph{Fine-tuned, full (FTF) \& Fine-tuned, decoder-only (FTD)}
Using the trained, task-omitted models discussed above, we can fine-tune these models on the \emph{omitted} task $t$ with loss $\mathcal L_t$ directly, allowing either the full (encoder \& decoder) model (\textbf{FTF}) or only the decoder (\textbf{FTD}) to update in fine-tuning. This style of training in which we pre-train on all tasks except for $t$, then fine-tune on task $t$ alone, is meant to simulate how we could adapt a pre-trained multi-task model to a novel task.

\subsection{Fine-tuning Settings}
\begin{figure}
    \centering
    \includegraphics[width=\linewidth]{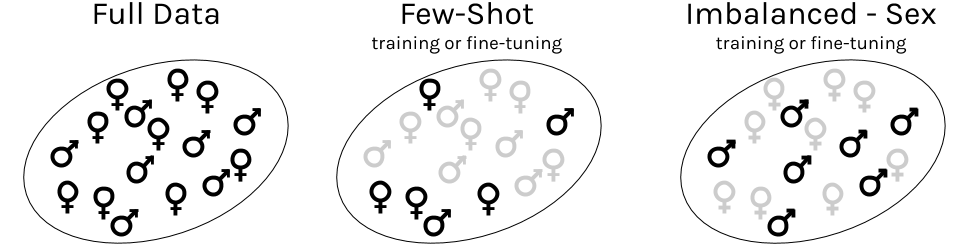}
    \caption{We assess models (either trained or fine-tuned) on three regimes: \emph{full-data} (left), \emph{few-shot} (middle), in which fine-tuning data was randomly subsampled to various degrees, and \emph{imbalanced} (right), in which fine-tuning data was subsampled in a sex-imbalanced fashion, with random percentages of genotypically female patients removed to simulate adapting to an imbalanced dataset.}
    \label{fig:fine-tuning_settings}
\end{figure}

We fine-tune our models under several modified versions of our data, shown graphically in Figure~\ref{fig:fine-tuning_settings}. These are designed to assess performance and to determine if MT pre-training can help us adapt models to few-shot or imbalanced datasets. First, we fine-tuned models over the full dataset in our \textbf{full-data} mode. Next, to simulate the adaptation of a pre-trained model to a new or rare disease with minimal data available, we fine-tuned ST, FTF, and FTD models in a \textbf{few-shot} setting, with the training data sub-sampled to various degrees.
Finally, to simulate adapting a model to a disease for which only imbalanced data is available in our \textbf{imbalanced} setting, we randomly subsampled female patients (by genotypical sex~\cite{johnson_mimic-iii_2016}) within the fine-tuning data to varying degrees, including total removal. 
In both the imbalanced and few-shot modes, note that pre-training is performed as described in Section~\ref{subsec:pre-training} on all tasks but the fine-tuning task $t$, over all available data, unaltered. Single task models used in these settings are naturally not pre-trained, and instead trained directly from scratch on the reduced datasets.
% Note that pre-trained models were still trained over the full data, but, as specified in Section~\ref{subsec:pre-training}, they were only trained over all tasks except the fine-tuning task $t$, so there is no leakage in this way.

% \paragraph{Full-Data} First, on all model types (ST, MT, FTF, and FTD), we fine-tuned models over the full dataset.

% \paragraph{Few-Shot}
% To simulate the adaptation of a pre-trained model to a new or rare disease with minimal data available, we trained/fine-tuned\footnote{Multi-task pre-training data was unchanged, but did not include fine-tuning tasks.} ST, FTF, and FTD models in a \emph{few-shot} setting, with the training data sub-sampled to various degrees.

% \paragraph{Imbalanced}
% To simulate adapting a model to a disease for which only imbalanced data is available, we randomly subsampled female patients (by genotypical sex~\cite{johnson_mimic-iii_2016}) within the training data to varying degrees, including total removal.

\section{Experimental Settings}
\label{sec:experiments}

\paragraph{Data}
%shorten this
We use the MIMIC-III dataset, containing intensive care unit (ICU) visits from patients in the Beth Israel Deaconess Medical Center between 2001-2012 \cite{johnson_mimic-iii_2016}. Data is parsed using the MIMIC-Extract \cite{wang_mimic-extract:_2019} pipeline and cohort under default parameters. We use the extracted time-series of labs and vitals, along with continuous embeddings of treatments as our input data and for our local auto-regressive tasks, and additionally use the static outputs and ICD codes for our tasks.
We prepare additional data out of MIMIC for our novel tasks, including 30-day readmission flags per-patient, and do not resuscitate (DNR) \& comfort measures only (CMO) code changes.\footnote{Full code and pre-processed data are available at \url{https://github.com/mmcdermott/comprehensive_MTL_EHR}.} We randomly split our data by patient via an 80-10-10 train, tuning, and test split, resulting in 17,500 patients in the train set, and 2188 patients in both the tuning and test sets. Minibatches are generated by sampling a random 48h sequence from a patient's EHR record each training epoch. Tasks deemed as \textit{rolling} and \textit{autoregressive} (See Table~\ref{tab:task_types}) are evaluated at 10 random time-points per patient. \emph{Static} tasks are evaluated over the first 24 hours per patient, and \emph{terminal} tasks are evaluated over the 48 hours prior to discharge/death.

\paragraph{Tasks}
%fts could use continuous dose Estimating the Effects of Continuous-valued Interventions using Generative Adversarial Networks mihaelia

\label{sec:tasks}
We use several tasks to assess the efficacy of MTL. We group these tasks into ten categories to ensure omitting a task group removes all highly correlated tasks. Scores are reported at a per-category level, averaging macro AUROCs across all tasks within that category. Full descriptions of the tasks can be found in the supplementary material~\ref{subsec:task_appendix}.

\newcommand{\mtr}{\multirow[t]{2}{*}}
\begin{table}
    \scriptsize
    \centering
    \caption{
    Tasks that are used in the multi-task learning ensemble. Majority class accuracy (MCA) is reported for all classification tasks (macro, if task is multi-label) to give an estimate of the relative level of class imbalance of the task. Number of observed labels is reported for all tasks, reflecting the differences both between static and dynamic tasks (the former having one label per patient, the latter one per patient per hour), as well as reflecting that some tasks only have valid labels on a subset of patients/patient-hours. Both Future Treatment Sequence (FTS), and our more granular Final Acuity (ACU) task are, to the best of our knowledge, novel tasks.
    \\ \textit{Abbreviations:} \emph{AR}: Autoregressive, \emph{Bin.}: binary classification, \emph{ML:} binary multi-label classification, \emph{MC:} multi-class classification, \emph{SMC}: sequential decoding multi-class classification, \emph{Reg.}: regression.
    }
    \label{tab:task_types}

    \begin{tabular}{p{0.15\textwidth}llllllllrr} \toprule
        Task Category              & Abbr.       & Specific Task                      & Temporal & Gap & Pred. & Type & Rel. Work  & \# Labels & MCA \\ %& Prior Art \\
        \midrule
        \mtr{Imminent Mortality}   &  \mtr{MOR}  & Mortality (24h)                    & Rolling  & 2h  & 24h   & Bin. & \mtr{\cite{harutyunyan_multitask_2017}}        &    1.8M       &   0.980 \\ 
                                   &             & Mortality (48h)                    & Rolling  & 6h  & 48h   & Bin. &            &     1.7M      &  0.962   \\ %& \\
        % \midrule
        \mtr{Comfort Measures}     &  \mtr{CMO}  & CMO added (24h)                    & Rolling  & 2h  & 24h   & Bin. & \mtr{\cite{lojun2010investigating}} &          1.8M &   0.992  \\ %& \multirow{2}{*}{} \\
                                   &             & CMO added (48h)                    & Rolling  & 6h  & 48h   & Bin. & &        1.7M   &  0.987   \\ %& \\
        % \midrule
        \mtr{DNR Ordered}          &  \mtr{DNR}  & DNR added (24h)                    & Rolling  & 2h  & 24h   & Bin. & \mtr{\cite{lojun2010investigating}} &      1.6M     &  0.988   \\ %& \multirow{2}{*}{} \\
                                   &             & DNR added (48h)                    & Rolling  & 6h  & 48h   & Bin. & &      1.6M     &  0.981   \\ %& \\
        % \midrule
        \mtr{Imminent Discharge}   &  \mtr{DIS}  & Discharge (24h)                    & Rolling  & 2h  & 24h   & MC   & \mtr{\cite{bertsimas2020predicting}} &      1.5M    &  0.730  \\ %&           \\
                                   &             & Discharge (48h)                    & Rolling  & 6h  & 48h   & MC   &        &   1.4M        &  0.473  \\ %&           \\
        % \midrule
        ICD Code Prediction        &  ICD        & See Suppl. Mat.~\ref{subsec:task_appendix_details}  & Static   & 12h & N/A   & ML   & \cite{ding2019effectiveness, harutyunyan_multitask_2017}      &     7.7K      &  0.691 \\ %&           \\
        % \midrule
        Long Length-of-Stay        &  LOS        &                                    & Static   & 12h & N/A   & Bin. & \cite{pmlr-v106-nestor19a, harutyunyan_multitask_2017, wang_mimic-extract:_2019}         &      21.9K     &   0.529 \\ %&           \\
        % \midrule
        30 Day ICU\\Readmission    &  REA        &                                    & Terminal & N/A & N/A   & Bin. & \cite{harutyunyan_multitask_2017}        &      21.9K     & 0.950   \\ %&           \\
        % \midrule
        Final Acuity               &  ACU        &                                    & Static   & 12h & N/A   & MC   & \cite{Che2018,wang_mimic-extract:_2019, Si2019} &      21.9K     &  0.253  \\ %&           \\
        % \midrule
        \mtr{Next Timepoint}       &  \mtr{WBM}  & Will-be-measured                   & AR       & 0h  & 1h    & ML   & \mtr{\cite{chang19a}} &     1.8M      & 0.920   \\ %&           \\
                                   &             & Next hour                          & AR       & 0h  & 1h    & Reg. &  &      1.8M     &  N/A  \\ %&           \\
        % \midrule
        Future Treatment\\Sequence &  FTS        &                                    & AR       & N/A & N/A   & SMC  & \cite{Wu_vassopressur_2016, peng2018improving}&    1.8M    &  N/A  \\ %&           \\
    \bottomrule \end{tabular} 
\end{table}

In addition to conventional classification tasks, we have two autoregressive tasks in our ensemble: multilabel prediction of which labs/vitals will be measured in the next hour, and continuous regression to the labs/vitals observed in the next time-point (presuming measured). Next hour regression was included in our ensemble as it was observed to be helpful to other tasks in preliminary results, but we do not report scores on the regression task as there is no clear analog to classification performance metrics and, despite the fact that this task was helpful on other tasks in training, all tested models' overall performance on this task were no better than mean imputation.

We also use a novel \emph{future treatment sequence} (FTS) task, for which individual treatments are first aggregated into categories (\emph{ventilation}, \emph{vasopressors}, and \emph{fluids}), then formed into a sequential decoding task in which the model predicts which subsets of treatments are used over the remainder of the patient's stay in sequence, but ignoring the duration of application of those treatments.
% in a duration agnostic manner where we predict simply the unique sequence of treatments seen, not the duration for which those treatments were applied.
% ~\cite{10.1093/jamia/ocu051}.
While all of our other tasks use a single fully connected layer for their task decoders, this sequential task uses a LSTM recurrent neural network with teacher forcing~\cite{lamb2016professor}.

\paragraph{Model Architecture}
Data is first projected into a common embedding space via a linear layer, then passed into a sequential encoder to produce a fixed size representation. Finally, this fixed-size representation is fed into the task-specific prediction heads (a single fully connected layer with a \texttt{softmax} activation, except for FTS) to yield the task-specific output (See Figure~\ref{subfig:model}). We investigated a linear, gated recurrent unit (GRU) recurrent neural network model~\cite{cho-etal-2014-learning}, and transformer architecture~\cite{vaswani_attention_2017} for our shared encoder; however the main body of the paper will feature only the GRU results for brevity, as this was our best performing model and it (or variants thereof) is well-established in this space~\cite{che2018recurrent}. Full details and results for all architectures are reported in the supplementary materials. To assess significance and give variance estimates, we re-trained our models over the same train/validation/test split with different random initialization.\footnote{Other schemes would require re-tuning hyperparameters, which was computationally infeasible.}

\paragraph{Hyperparameter Tuning}
%here or in methods
Hyperparameter tuning was performed using the \texttt{Hyperopt} library~\cite{bergstra_making_2013}, optimizing for the average AUROC across all tasks in the multi-task setting. These hyperparameters were applied to ST, FTD, and FTF models. Multi-task training was used for hyperparameter tuning like this for computational efficiency; if we performed a separate hyperparameter tuning run for all tasks independently, it would inflate the hyperparameter search time by a factor of 10, which was not feasible. To assess if choosing hyperparameters based solely on MTL performance introduced bias, we ran a single task through a separate hyperparameter tuning run, and analyzed the results of our main run as though we were choosing to optimize a single task (for example, instead of summing the losses of 10 tasks, only 1 task loss is used, which which may have required a higher learning rate to converge as fast as the MT models). In both cases, the difference in ultimate performance was negligible, so we concluded that this bias was an acceptable risk given the extent of the computational savings it offered. This effect would bias us in favour of pure multi-task (MT) results, whereas we do not find these to be top performing in practice. Further discussion of the hyperparameter search can be found in Supplementary Material~\ref{sec:hyperparameter_details}. 

\section{Results \& Discussion}
\label{sec:results_discussion}

In this section, we report results and comment on the GRU system across our four guiding questions. Full results for all model types are present in the supplementary material, Section~\ref{sec:all_arch_results}. 

\subsection{Most Tasks Have Negative Transfer for MTL}
\label{subsec:neg_transfer}
\begin{figure}
    \centering
    \includegraphics[width=\linewidth]{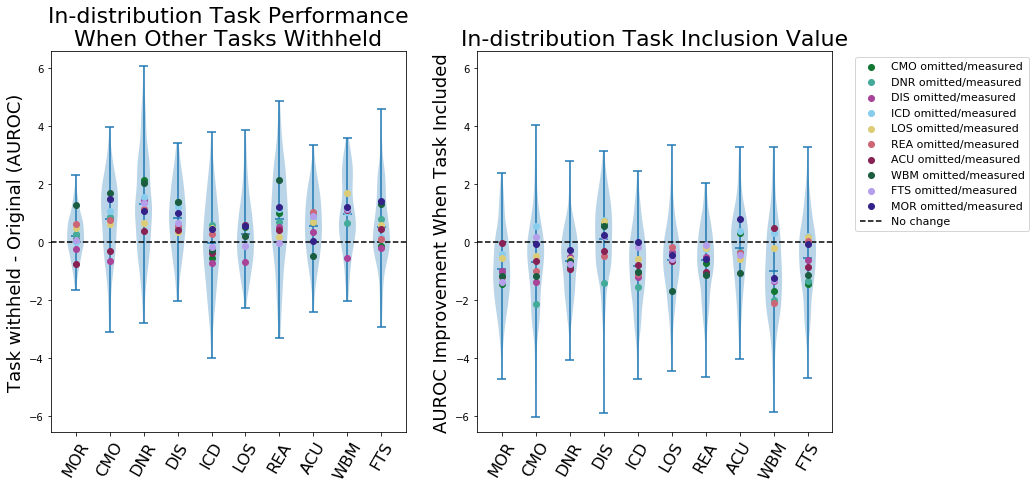}
    \caption{
    Performance (AUROC, scaled by 100) change on task $r$ between the full MT system $\mathcal M(r)$ and task $t$-omitted MT system $\mathcal M_{\neg t}(r)$ (recall Section~\ref{sec:training_regimes}), shown in 2 ways.
    % \textit{Left} The performance difference $\mathcal M_{\neg t} - \mathcal M$ ($y$-axis) of the system on each individual task ($x$-axis) as other tasks $t$ (colored dots) are removed.
    % \textit{Right} The performance difference $\mathcal M - \mathcal M_{\neg t}$ ($y$-axis) on each individual tasks (colored dots) when including each auxiliary task $t$ ($x$-axis). \\
    % Note that the two plots show the same values (they are simply inverted on the right), but organized differently --- 
    On the left, we show $\mathcal M_{\neg t}(r) - \mathcal M(r)$ ($y$-axis) to demonstrate how performance on each individual task $r$ ($x$-axis) changes as other tasks $t$ (colored dots) are removed.
    On the right, we show the same values (though inverted, with $\mathcal M(r) - \mathcal M_{\neg t}(r)$ on the $y$-axis), but transposed: Now task $t$ is on the $x$-axis, showing how including $t$ affects the performance on all other tasks $r$ in the ensemble (now via colored dots).
    \\
    Each dot represents the mean of the relevant difference taken over several random samples - the width of the violin plot reflects the distribution of all possible differences across all tasks.
    \\
    We can see from the left plot that on almost all tasks, $\mathcal M_{\neg t}(r)$ is larger than $\mathcal M(r)$, indicating significant negative transfer.
    Similarly, on the right we can see that there are no ``universally'' helpful tasks to include and some tasks are consistently harmful.
    }
    \label{fig:neg_transfer_gru}
\end{figure}

By examining our intermediate, task-omitted pre-training results, we can assess the propensity of negative transfer directly. In Figure~\ref{fig:neg_transfer_gru}, we can see that the performance of our models when a single task is withheld is often greater than their performance in the full multi-task regime, indicated by the majority of the mass of the violin-plots on the left figure being greater than zero. We can also examine these results as a function of which tasks are being included in the ensemble from our right plot, seeing that 
% we plot the difference in observed AUROC between a full MT system and a MT system with a single task omitted ($\mathcal M - \mathcal M_{\neg t}$, as described in Section~\ref{sec:training_regimes}) in two ways. We can see that most of the mass of the violin-plots in this graph is greater than zero for in-distribution performance when other tasks are withheld. This indicates that performance largely \emph{improves} when other tasks are removed. This allows us to assess whether any tasks selectively perform better in a MT regime.
% We also see from the In-distribution Task Inclusion Value plot that
there is no single task which, when added to the distribution, offers a consistent improvement on other tasks. This corroborates empirical and theoretical evidence that only highly correlated tasks result in positive task transfer~\cite{ndirango2019generalization, Wu2020Understanding}.
In general, both views support that task performance under direct MT training is generally hurt when more tasks are included in the ensemble, and there are no clear outlier tasks that consistently improve performance when included, indicating that negative transfer is prevalent.

\subsection{Pre-training, Followed by FTF or FTD Fine-tuning Improves Final Performance}
\label{sec:main_results}
The final performance of the GRU model across all 4 main MTL regimes (ST, MT, FTD, and FTF) in the full data setting is reported in Table~\ref{tab:gru_results}. There are several takeaways from this table. First, some variant of multi-task training reaches best-in-class performance on all but ICD Code Prediction, and some variant of fine-tuning performs best on all remaining tasks except final acuity prediction (ACU). Additionally, note that FTF alone outperforms or matches ST performance on all tasks but ICD code and imminent discharge (DIS) prediction. In the full-data setting only DNR and WBM show statistically significant comparisons between ST and FTF, and REA and WBM show statistically significant comparisons between ST and FTD, in all cases with fine-tuning performing better (significance assessed via a $t$-test at $p < 0.05$, $n=5$). Overall this indicates that fine-tuning is capable both of matching original ST performance certainly, and possibly extending beyond it across this wide range of tasks. Note that this is true \emph{despite} the fact that MT performance underperforms ST performance on all tasks save 30-day readmission (REA), ACU, and the will-be-measured task (WBM). This suggests that the fine-tuning is synergistically building on the strengths of both MT and ST training.
Second, FTF outperforms or matches FTD results on all tasks DIS, REA, and WBM. This suggests that the regularization benefit of freezing the encoder layers can be helpful, but as a general rule one \emph{should} fine-tune the full architecture rather than just the decoder.
Third, the fact that MT consistently underperforms ST training underscores our findings from Section~\ref{subsec:neg_transfer}, suggesting that negative transfer is common here.

\begin{table}
    \centering
    \scriptsize
    
    \caption{GRU Results (AUROC, scaled by 100) subdivided among different MTL regimes, under both the full-data fine-tuning setting and few-shot (1\%) setting. Bolded results indicate top performing result per each task/evaluation setting. We can see that on all tasks save ICD code prediction, some variant of multi-tasking, most commonly FTF or FTD,  matches or improves over ST training by some margin in the full-data setting, and on all tasks save DIS \& FTS, some variant of fine-tuning outperforms ST training in the few-shot setting, sometimes by impressive margins.}
    \label{tab:gru_results}

    \begin{tabular}{llllllll}
    \toprule
    \multirow[c]{2}{*}{Task} & \multicolumn{4}{l}{Full-data} & \multicolumn{3}{l}{Few-shot (1\%)} \\
    \cmidrule(r{4pt}){2-5} \cmidrule(l){6-8}
    {}                       &                           ST  &                           MT &                          FTD &                          FTF &                           ST &                          FTD &                          FTF \\
    \midrule
    MOR                      &               $94.5 \pm 0.5$ &               $94.1 \pm 0.5$ &               $94.4 \pm 0.4$ &  $\boldsymbol{94.9 \pm 0.2}$ &               $68.4 \pm 7.6$ &              $39.6 \pm 12.2$ &  $\boldsymbol{89.4 \pm 2.2}$ \\
    CMO                      &               $91.3 \pm 0.7$ &               $90.5 \pm 1.0$ &               $91.4 \pm 1.0$ &  $\boldsymbol{92.3 \pm 0.9}$ &               $50.5 \pm 5.9$ &              $42.0 \pm 15.7$ &  $\boldsymbol{75.4 \pm 7.6}$ \\
    DNR                      &               $88.0 \pm 1.0$ &               $87.2 \pm 1.5$ &               $88.9 \pm 0.7$ &  $\boldsymbol{90.2 \pm 0.9}$ &               $54.1 \pm 8.9$ &               $30.7 \pm 8.6$ &  $\boldsymbol{76.3 \pm 1.3}$ \\
    DIS                      &               $78.8 \pm 1.6$ &               $78.3 \pm 0.9$ &  $\boldsymbol{79.0 \pm 0.3}$ &               $78.6 \pm 2.0$ &  $\boldsymbol{57.7 \pm 4.1}$ &               $50.7 \pm 0.6$ &               $57.3 \pm 0.7$ \\
    ICD                      &  $\boldsymbol{71.1 \pm 0.7}$ &               $69.2 \pm 1.0$ &               $70.6 \pm 1.5$ &               $70.9 \pm 3.1$ &               $53.2 \pm 2.6$ &               $51.1 \pm 2.5$ &  $\boldsymbol{56.7 \pm 1.1}$ \\
    LOS                      &               $72.8 \pm 0.5$ &               $71.6 \pm 0.9$ &               $72.8 \pm 0.2$ &  $\boldsymbol{72.9 \pm 1.0}$ &               $52.4 \pm 0.6$ &              $53.3 \pm 11.7$ &  $\boldsymbol{66.4 \pm 3.2}$ \\
    REA                      &               $58.4 \pm 1.5$ &               $60.9 \pm 1.4$ &  $\boldsymbol{62.6 \pm 0.7}$ &               $59.3 \pm 3.8$ &               $51.7 \pm 4.5$ &  $\boldsymbol{51.8 \pm 6.3}$ &               $51.8 \pm 4.5$ \\
    ACU                      &               $78.7 \pm 1.0$ &  $\boldsymbol{80.1 \pm 0.9}$ &               $79.9 \pm 0.5$ &               $80.0 \pm 2.7$ &               $56.1 \pm 2.7$ &               $52.4 \pm 4.9$ &  $\boldsymbol{60.9 \pm 1.1}$ \\
    WBM                      &               $71.8 \pm 0.5$ &               $73.8 \pm 1.1$ &  $\boldsymbol{76.8 \pm 0.4}$ &               $73.5 \pm 1.4$ &               $50.6 \pm 0.1$ &               $50.2 \pm 0.8$ &  $\boldsymbol{51.1 \pm 1.8}$ \\
    FTS                      &               $89.2 \pm 1.2$ &               $87.9 \pm 1.0$ &               $88.3 \pm 0.6$ &  $\boldsymbol{90.1 \pm 0.6}$ &  $\boldsymbol{63.6 \pm 3.9}$ &               $55.5 \pm 6.1$ &               $58.8 \pm 2.0$ \\
    \bottomrule
    \end{tabular}
\end{table}

\subsection{FTF Greatly Improves Few-Shot Learning Performance}
\begin{figure}
    \centering
    % \begin{subfigure}{\linewidth}
    %     \centering
    %     \includegraphics[width=\linewidth]{Images/small_data_gru_all_fracs_mor_los_wbm.png}
    %     % \caption{AUROC (of 100; $y$-axis) of 3 representative tasks (plot; line color) of FTF, FTD, and ST models (line style) across various subsampled training/fine-tuning datasets ($x$-axis). Higher is better. we can see that TODO.}
    %     % \label{subfig:raw_nums}
    % \end{subfigure}
    % \begin{subfigure}{\linewidth}
    %     \centering
    %     \includegraphics[width=\linewidth]{Images/small_data_gru_vs_baseline_all_fracs_grouped.png}
    %     % \caption{AUROC (of 100; $y$-axis) difference between MTL regime (line style) and a ST model of all tasks (line color) broken down into rolling tasks (left), static tasks (middle), and autoregressive tasks (right) across various subsampled training/fine-tuning datasets ($x$-axis). Higher is better. we can see that TODO.}
    %     % \label{subfig:raw_nums}
    % \end{subfigure}
    
    \includegraphics[width=\linewidth]{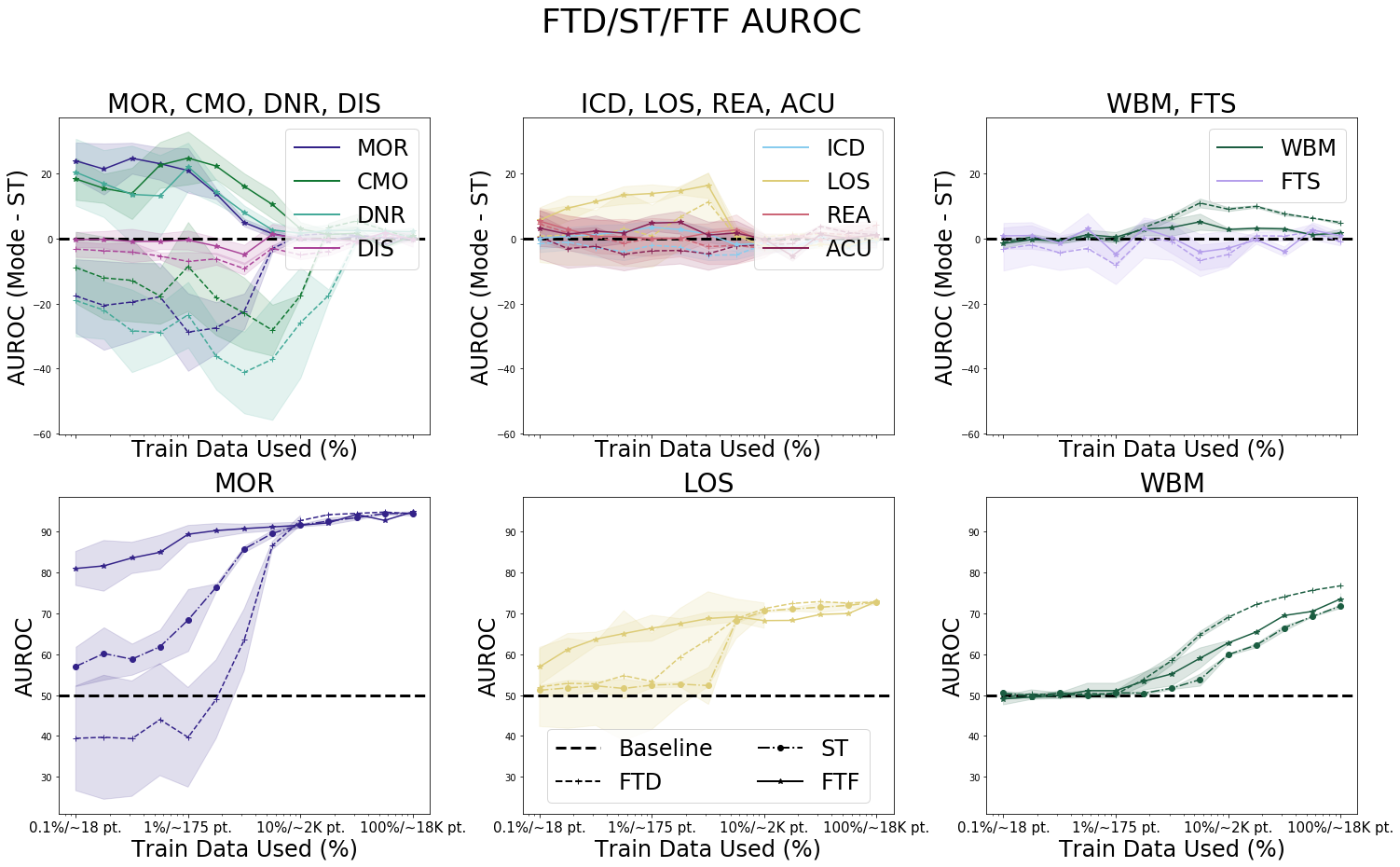}
    
    % \caption{AUROC, both raw (top) and compared to a ST baseline (bottom) of multiple tasks and training styles as a function of dataset subsampling rate ($x$-axis).}
    \caption{
    AUROC, both compared to a ST baseline (top) and raw (bottom) of multiple tasks (line color) and training styles (line style) as a function of dataset subsampling rate ($x$-axis). Higher is better. Tasks in the upper plots are grouped according to rolling tasks (left), static tasks (middle), and autoregressive tasks (right), with a particular example highlighted in raw units across all three training styles above. We can see that on our rolling tasks, FTF models tend to perform much better in general even at drastically smaller dataset subsampling rates than ST models, and FTF training for mortality retains strong performance even for only 0.1\% of the data.
    }
    \label{fig:small_data_all_fracs}
\end{figure}

%1 sentence taskeaway
Final performance of each of our pre-training regimes and our ST baseline on various levels of reduced fine-tuning/training data (simulating increasingly limited few-shot contexts) is shown in Figure~\ref{fig:small_data_all_fracs}, with a 1\% dataset size highlighted explicitly in Table~\ref{tab:gru_results}. On all tasks except FTS, DIS, and REA, FTF offers improvements over both FTD and ST training even in the extreme levels of dataset reduction, by margins ranging up to approximately 25\% improvements over ST training. The comparisons at the 1\% data level between FTF and ST for MOR, CMO, DNR, LOS, and ACU are all statistically significant (computed via a $t$-test, at $p < 0.05$); notably, this means both instances where ST outperforms FTF (DIS and FTS) are \emph{not} statistically significant deviations.
We can see that these gains are particularly dramatic for the MOR, CMO, DNR, and LOS task, which retain strong improvements over ST training even to as low as 0.1\% dataset size, when only approximately 20 patients would be used for fine-tuning. For MOR, performance under FTF drops only by approximately 10\% between full-data and 0.1\% data, whereas for ST training it drops by nearly 30\%. Note that we consistently see the largest gains on our rolling, binary classification tasks (MOR, CMO, and DNR), all of which show significant class imbalance (Table~\ref{tab:task_types}). This may suggest that this strategy is particularly suited to rolling tasks, imbalanced tasks, or binary classification tasks.

While these results do align with prior results in the natural imaging domain that multi-task pre-training can lead to significant advances in the few-shot domain~\cite{tian2020rethinking}, surprisingly FTD consistently underperforms FTF training. We expected that, given the reduced training set size, the decreased capacity enforced by a frozen encoder would have given FTD training an edge over FTF training, but this does not appear to be the case; instead, allowing the full model to tune is essential. In the context of clinical data, where diverse rare diseases are prevalent~\cite{wakap2020estimating, 10.1001/jamanetworkopen.2020.1965} and, even in the context of well understood diseases, individual treatment trajectories can be highly specialized~\cite{hripcsak2016characterizing}, the ability to fine-tune successfully on so little data is very valuable.

\subsection{MTL Pre-training Does Not Address Minority Class Imbalance}
We do not find that MTL pre-training is able to reduce performance bias in favor of the majority group in our experiments. While this does not eliminate the possibility that some form of MTL can aid in reducing bias, as others have reported in other domains, from our work there is no evidence to suggest that multi-task pre-training on a balanced dataset, followed by task-specific fine-tuning on an imbalanced dataset reduces model bias over single-task training on the imbalanced dataset directly. We should note, however, that most discrepancies observed in our experiments were small overall, even when all female patients were removed entirely from the fine-tuning dataset, so it may be that this approach would still offer some gains under datasets/models with greater degrees of bias. 
% Race influences clinical outcomes~\cite{martin2011color}. The validity of this stratification has recently been questioned by the clinical community, and is still being explored \cite{eneanya2019reconsidering, NIPS2018_7613}.
We present full commentary on these results in Supplementary Material Section~\ref{sec:fairness}.

% Fortunately performance discrepancy isn't consistently observed even when severely tipping the favour away from the protected group. There is no evidence to suggest that you can make a task fairer by pre-training on a fair set of tasks. Perhaps this effect may be observed in clinical tasks with larger disparities~\cite{NIPS2018_7613}. 

% Future work could investigate sensitive groups for unfairness under FTF and especially few-shot learning schemes. The addition of data from other sources, will provide unique demographics with would bolster our confidence that we are doing equitable MTL.
% Lastly, we also performed clustering analyses of the induced representations of the MT pre-training system, to see if we detect notable subclusters or an improved ability to subtype patients. This was not observed. Full details of the fairness and clustering studies are available in Supplementary Material Sections~\ref{sec:fairness} and \ref{sec:clustering}, respectively.

\section{Conclusion}
% \paragraph{Future Work}
% While this work offers an extensive analysis of the efficacy of MTL on EHR data, there are several important directions for Future Work.
% First, repeats of this analysis under different datasets on healthcare domains would help clarify the generalizability of these findings outside of MIMIC-III. Our tasks could also be improved in future work with the inclusion of other self-supervised training strategies for time-series data~\cite{oord2018representation, NIPS2019_8568}, or improvements to how we work with and represent more complex task types. Lastly, additional investigation is warranted into the opportunities to use MTL to improve modeling over imbalanced datasets; while our preliminary results suggest minimal improvement, further work is necessary to fully understand the opportunities afforded by MTL in this context.

% \paragraph{Key Takeaways}
In this work, we defined a battery of tasks over EHR physiological time-series data, including two novel tasks, and used these tasks to profile a battery of MTL strategies on clinical data. We find that while using traditional MTL results in systemic negative transfer, using a MT, task-independent pre-training scheme, followed by task-specific fine-tuning, yields modest improvements under standard fine-tuning, and can yield dramatic improvements in the few-shot context (up to a gain in AUROC of approximately 25\% on 1\% training data). This approach allowed our model to retain nearly maximal performance on prediction of imminent mortality with as little as 1\% training data.
We do not, however, find consistent evidence to suggest that FTF is more or less susceptible to pitfalls of training on population imbalanced data.
This paper suggests that while MTL pre-training may not ameliorate model bias on imbalanced datasets, it nonetheless does offer a scalable vehicle for improving performance in important clinical settings, including rare disease detection.

% This could include bypass networks \todo{I don't know what this is}, soft layer sharing~\cite{duong-etal-2015-low}, curriculum learning~\cite{pentina2015curriculum}, ensemble architectures, gradient normalization \cite{pmlr-v80-chen18a}, etc. \todo{citations} 
%Some of these approaches naturally incorporate a vehicle to trade off single-task learning and multi-task learning during training, thereby avoiding negative transfer and potentially approaching the performance of our FTF settings. 

\section*{Broader Impact}
% This section is required in NeurIPS 2020.
This paper explores the utility of multi-task learning to solve foundational challenges in machine learning for health and biomedicine. Of particular interest in our analyses is the ability of multi-task learning to aid in generalizing to smaller datasets, to simulate an application of a model to a rare disease setting, and to underrepresented subgroups, to assess the utility of multi-task learning to improve fairness concerns. While we find that this work does not offer improvements to fairness, and thus cannot be used to help solve that problem, we do find significant improvements in the few-shot setting which could aid modelling of rare or emerging diseases, or where large-scale labelling is expensive or invasive. Further, the knowledge that this technique does not, at present, appear to help ameliorate bias, helps the field understand the complex challenges faced to ensure our models are fair and help all patients. 

\bibliographystyle{plain}
\bibliography{main}

\clearpage

\appendix

% \todo{exact cohort \#s for all experiments (few-shot, gender bias, race bias).}

\section{Detailed Task Definitions}
\label{subsec:task_appendix}
\subsection{Task Selection Overview}
% \todo{this subsection}

We intentionally chose a diverse set of tasks designed to span the kinds of tasks we often assess in ML4H. These include clinically motivated tasks, such as those that measure acuity, likelihood of future treatments, or which labs \& vitals will be measured in the next time-point. Operationally motivated tasks may be included, such as prediction of imminent discharge, ICD billing codes, and readmission risk. We also include a local auto-regressive task to encourage the model to maintain a faithful, full capacity representation of the input. 

The goal of selecting such a diverse battery of tasks, even while common MTL literature suggests that tasks must be similar in order for MTL to offer benefits~\cite{ruder2017overview}, is that in a pre-training/fine-tuning context, in which we ostensibly \emph{don't know} what the fine-tuning task will look like at the time of pre-training, we need the diverse ensemble to ensure generalizability to a wide range of tasks. Further work could examine under what arrangements of task similarity MT pre-training offers benefits on fine-tuning tasks. However, we intuit that pre-training on a diverse set of tasks, rather than a tightly correlated set (which may be very divergent from the fine-tuning task), would offer the best performance on an unknown downstream task in the context of a pre-training/fine-tuning regime.

Below, we detail all our tasks used, their precise source in the input data, related work on each task, and give baseline statistics about their label spaces and chance frequencies.

\subsection{Our Tasks}
\label{subsec:task_appendix_details}

% \todo{So far we have two new citations in here! One is likely rajkomarscalable...}

% We predict four imminent acuity event tasks, designed to serve as proxies for real-time early warning systems detecting upcoming decompensation events: prediction of mortality~\cite{Che2018, PMID:29893864, harutyunyan_multitask_2017}, discharge, CMO code changes, and DNR/DNI code changes, each over both a 24-hour/2-hour and 48-hour/6-hour prediction/gap window.

% We also predict a final acuity event task which extends the common in-hospital mortality prediction task by covering a more granular target space, including mortality of various kinds and discharge to various locations (full breakdown of labels in Supplementary Material Section~\ref{subsec:acuity_appendix}).

% We also include three commonly studied non-time-varying tasks which could help provide operational support to a clinic: ICD code prediction~\cite{ding2019effectiveness, PMID:29893864, harutyunyan_multitask_2017}, 30-day readmission prediction \cite{PMID:29893864, harutyunyan_multitask_2017, wang_mimic-extract:_2019}, and long length-of-stay (LOS) prediction~\cite{pmlr-v106-nestor19a, harutyunyan_multitask_2017, wang_mimic-extract:_2019}. ICD code prediction is done in a multi-label manner predicting the presence of any of 18 major ICD categories (full list in Supplementary Material Section~\ref{subsec:icd_app}). Long LOS is defined as a binary prediction task, differentiating between stays under or over 3-days (the approximate average LOS in our cohort).

\paragraph{Imminent Mortality: MOR} \hfill
\\ \textbf{Description:} We predict imminent mortality across both a 24h and 48h window, using 2h and 6h gap times, respectively. These predictions can be used as indicators of imminent physiological decompensation, and spanning multiple prediction windows gives the system incentive to learn a representation both reflecting immediate and urgent, but not necessarily immediate, signals of decompensation. This task is a binary classification task.
\\ \textbf{Data Source:} Time of death was extracted from MIMIC Extract's provided static output~\cite{wang_mimic-extract:_2019}. 
% \todo{validate that MIMIC-Extract resolves the deathtime v. outtime discrepancy, or comment on which we use.}
\\ \textbf{Prior Art:} Imminent mortality has been used as a proxy for physiological decompensation historically in several studies. Harutyunyan et al.~\cite{harutyunyan_multitask_2017}, for example, explore this task.% \todo{likely should add other examples}
\\ \textbf{Statistics:} Imminent mortality prediction is a highly imbalanced task, with approximately 98\%, 96.2\% of patients not dying within 24, 48 hours, respectively.

\paragraph{Comfort Measures: CMO} \hfill
\\ \textbf{Description:} ``Comfort Measures Only'' (CMO) orders indicate that the (usually terminally ill) patient has requested to receive care \emph{only} designed to provide comfort, not treatment, and otherwise the course of illness should be allowed to progress (typically to mortality).
% \todo{likely need citation for def'n of CMO} 
Predicting that a patient will soon add a CMO order provides another view towards a measure of imminent acuity. Like mortality, we predict CMO across both a 24h and 48h window, using a 2h/6h gap time, respectively. This task is a binary classification task.
\\ \textbf{Data Source:} This signal was extracted from MIMIC directly, via the \texttt{code\_status} table. A forked version of MIMIC Extract~\cite{wang_mimic-extract:_2019} with this and all other additions we made for our extraction will be made publicly available.
\\ \textbf{Prior Art:} In the traditional ML4H community, CMO prediction is somewhat understudied. However, examples do exist, such as in the work of Lojun et al.~\cite{lojun2010investigating}, who use natural language processing over clinical notes and structured data to predict CMO codes and do not resuscitate (DNR) codes.
\\ \textbf{Statistics:} CMO prediction is a highly imbalanced task, with roughly 99.2, 98.7\% of patients not registering a CMO order within 24, 48 hours, respectively.

\paragraph{DNR Ordered: DNR} \hfill
\\ \textbf{Description:} ``Do Not Resuscitate'' (DNR) orders indicate that the patient has requested to not receive resuscitation care (e.g., cardiopulmonary resuscitation a.k.a. CPR) and that, should those interventions be necessary, the patient should instead be allowed to die.
% \todo{likely need citation for definition of DNR}
Predicting that a patient will soon request a DNR order provides another view towards a measure of imminent acuity. Like mortality, we predict DNR across both a 24h and 48h window, using 2h and 6h gap times, respectively. This task is a binary classification task.
\\ \textbf{Data Source:} This signal was extracted from MIMIC directly, via the \texttt{code\_status} table. A forked version of MIMIC Extract~\cite{wang_mimic-extract:_2019} with this and all other additions we made for our extraction will be made publicly available.
\\ \textbf{Prior Art:} In the traditional ML4H community, CMO prediction is somewhat understudied. However, examples do exist, such as in the work of Lojun et al.~\cite{lojun2010investigating}, who use natural language processing over clinical notes and structured data to predict CMO codes and do not resuscitate (DNR) codes.
\\ \textbf{Statistics:} DNR prediction is a highly imbalanced task, with roughly 98.8, 98.1\% of patients not yielding a new DNR order within 24, 48 hours, respectively.

\paragraph{Imminent Discharge: DIS} \hfill
\\ \textbf{Description:} Like the prior tasks, we predict imminent discharge across both a 24h and 48h window, using a 2h/6h gap time. Unlike the prior tasks, the imminent discharge task is a multi-class classification task, forcing the model to predict to where the patient will be discharged, among several possible destinations outlined in Table~\ref{tab:imminent_discharge_stats}. Whereas the former tasks provide a view into acuity by predicting an imminent event that indicates a heightened acuity, predicting imminent discharge indicates that the patient's has become less acutely ill. Additionally, prediction of imminent discharge has operational benefits, by enabling hospitals to estimate how many beds will be free in the ICU in the near term future.
\\ \textbf{Data Source:} We use the discharge time and location provided in the MIMIC Extract static output~\cite{wang_mimic-extract:_2019}.
\\ \textbf{Prior Art:} Imminent discharge has been primarily predicted in operational contexts, rather than for use as a signal of acuity; for example, Bertsimas et al.~\cite{bertsimas2020predicting} predict imminent discharge to estimate patient flow and aid in scheduling.
\\ \textbf{Statistics:} Each possible discharge location, along with the percent of patients that are discharged to that location within 24 hours/48 hours of any given timepoint, respectively, are shown in Table~\ref{tab:imminent_discharge_stats}.

\begin{table}
    \centering
    \caption{All discharge locations we predict, along with the percent of patient-houts across the entire dataset that are discharged to that location within 24, 48 hours, respectively.}
    \begin{tabular}{lrr} \toprule
        Discharge Location                          & \% @ 24h & \% @ 48h \\ \midrule
        No Discharge                                & 73.0\%   & 47.3\%   \\
        Home Health Care                            & 7.7\%    & 15.1\%   \\
        Home                                        & 7.3\%    & 14.0\%   \\
        Skilled Nursing Facility (SNF)              & 5.2\%    & 10.3\%   \\
        Rehab/Distinct Part Hosp                    & 4.0\%    & 7.9\%    \\
        Long Term Care Hospital                     & 1.1\%    & 2.2\%    \\
        Discharge-Transfer Cancer/Children Hospital & 0.4\%    & 0.9\%    \\
        Short Term Hospital                         & 0.3\%    & 0.6\%    \\
        Discharge-Transfer To Psych Hospital        & 0.3\%    & 0.6\%    \\
        Hospice-Home                                & 0.3\%    & 0.5\%    \\
        Left Against Medical Advice                 & 0.1\%    & 0.2\%    \\
        Hospice-Medical Facility                    & 0.1\%    & 0.2\%    \\
        Home With Home Iv Provider                  & 0.0\%    & 0.1\%    \\
        Integrated Care Facility (ICF)              & 0.0\%    & 0.1\%    \\
        Other Facility                              & 0.0\%    & 0.1\%    \\
        Discharge-Transfer To Federal Hc            & 0.0\%    & 0.0\%    \\
        Snf-Medicaid Only Certif                    & 0.0\%    & 0.0\%    \\
    \bottomrule \end{tabular}
    \label{tab:imminent_discharge_stats}
\end{table}

\paragraph{ICD Code Prediction: ICD} \hfill
\\ \textbf{Description:}
We predict the multi-label presence of each major ICD category~\cite{slee1978international} (full list reported in Table~\ref{tab:icd_stats}). ICD code prediction can be seen as a phenotyping task; however, as ICD codes are applied to summarize a patient's entire stay for billing purposes, it is more accurately interpreted as an operational task which could aid a clinic's billing department. 
\\ \textbf{Data Source:} This is extracted from the provided default ICD code dataframe output by MIMIC Extract~\cite{wang_mimic-extract:_2019}.
\\ \textbf{Prior Art:} Prediction of ICD codes is commonly used as a phenotyping task~\cite{harutyunyan_multitask_2017,ding2019effectiveness}. 
% \cite{slee1978international} \todo{What's the deal with this citation}
\\ \textbf{Statistics:} 
Percentage of patients with at least one code in each ICD category used for this task are shown in Table~\ref{tab:icd_stats}. Recall that results are reported via macro AUROC across all labels, so though there are a subset with extreme class imbalance, they only play a moderate role on the overall AUROC reported for this task.

\begin{table}
    \centering
    \caption{Percent of patients who have at least one ICD code within each of the below categories.}
    \begin{tabular}{lr} \toprule
        Category        & \% Patients \\ \midrule
        Circulatory     & 72.2\% \\ 
        Endocrine       & 63.5\% \\
        Respiratory     & 53.0\% \\
        Injury          & 50.0\% \\
        Digestive       & 48.9\% \\
        Ill Defined     & 48.7\% \\
        Genitourinary   & 48.0\% \\
        Blood           & 47.9\% \\
        Mental Health   & 43.2\% \\
        Infection       & 41.8\% \\
        Nervous         & 40.8\% \\
        Musculoskeletal & 33.5\% \\
        Neoplasm        & 29.8\% \\
        Skin            & 22.7\% \\
        Congenital      & 8.3\% \\
        Pregnancy       & 1.2\% \\
        Unknown         & 0.02\% \\
        Perinatal       & 0.00\% \\
    \bottomrule \end{tabular}
    \label{tab:icd_stats}
\end{table}

\paragraph{Long Length-of-Stay: LOS} \hfill
\\ \textbf{Description:} We translate the "remaining LOS" regression task into a binary classification task by classifying patients as either below or above the average length of stay, rounded to the nearest day, in our cohort, which was 3 days. 
\\ \textbf{Data Source:} We use the default LOS output in MIMIC Extract's static outputs~\cite{wang_mimic-extract:_2019}.
\\ \textbf{Prior Art:} Long LOS has been predicted numerous times, both in a classification sense for 3-day LOS~\cite{wang_mimic-extract:_2019} and 7-day LOS~\cite{harutyunyan_multitask_2017}.
% It has also been predicted in a regression context \todo{citation}
\\ \textbf{Statistics:} This is a balanced prediction task, with a positive rate of approximately 52.9\%.

\paragraph{30 Day ICU Readmission: REA} \hfill
\\ \textbf{Description:} Rapid readmission is a serious operational concern to clinics, as they face financial penalties from certain insurance providers if a patient is discharged, but rapidly requires readmission. Given the limitations of the MIMIC-III data, which only covers data for patients admitted to the ICU, we predict solely 30-day ICU readmission, so that we can trust both our positive and negative labels, in a binary classification context. As MIMIC-Extract extracts a cohort only of patients' \emph{first} ICU stays~\cite{wang_mimic-extract:_2019}, this task also has the bias of only being analyzed on a new ICU visit for a patient, and would not be applicable for a population of repeat patients.
\\ \textbf{Data Source:} We constructed labels for this task our-self, by looking to see if there was a second record of an ICU admission for that patient within MIMIC within 30 days. We used MIMIC's \texttt{intime} and \texttt{outtime} as our definitive record of admission start and end times.
\\ \textbf{Prior Art:} Rajkomar et al.~\cite{rajkomar2018scalable} examine overall hospital readmission in their work.
% \todo{others}
\\ \textbf{Statistics:} This task is a relatively imbalanced task, with approximately 95\% of patients not being readmitted.

\paragraph{Final Acuity: ACU} \hfill
\\ \textbf{Description:}
This task is an extension of the common in-hospital mortality prediction task. The labels cover a more granular target space, outlined in Table~\ref{tab:acuity_stats_stats}. This extension makes the task much more difficult, and renders our results not directly comparable to previously published numbers.
\\ \textbf{Data Source:} We use the death time and discharge locations output by default in the MIMIC Extract static output for this task~\cite{wang_mimic-extract:_2019}.
\\ \textbf{Prior Art:} Various sub-forms of this task have been explored historically. In-ICU and in-hospital mortality, for example, have been explored in numerous ways~\cite{harutyunyan_multitask_2017,wang_mimic-extract:_2019,che_recurrent_2018}. Prediction of final discharge location is an extended version of the mortality task. Challenging the model to predict over both spaces jointly is novel, to the best of our knowledge.
\\ \textbf{Statistics:} Label values and the percent of patients with each label are shown in Table~\ref{tab:acuity_stats_stats}.

\begin{table}
    \centering
    \caption{Possible labels for our ``Final Acuity'' task, with the \% of Patients that have that label in our cohort.}
    \begin{tabular}{lr}\toprule
        Final Acuity Event                          & \% Patients \\ \midrule
        Discharge to Home Health Care               & 25.3\% \\
        Discharge to Home                           & 24.0\% \\
        Discharge to SNF                            & 17.2\% \\
        Discharge to Rehab/Distinct Part Hosp       & 13.2\% \\
        In ICU Mortality                            & 7.4\%  \\
        In Hospital Mortality                       & 3.7\%  \\
        Discharge to Long Term Care Hospital        & 3.6\%  \\
        Discharge-Transfer Cancer/Children Hospital & 1.5\%  \\
        Discharge to Short Term Hospital            & 1.1\%  \\
        Discharge-Transfer To Psych Hosp            & 1.0\%  \\
        Discharge to Hospice-Home                   & 0.9\%  \\
        Left Against Medical Advice                 & 0.4\%  \\
        Discharge to Hospice-Medical Facility       & 0.3\%  \\
        Discharge to Home With Home Iv Provider     & 0.2\%  \\
        Discharge to ICF                            & 0.1\%  \\
        Discharge to Other Facility                 & 0.1\%  \\
        Discharge-Transfer To Federal Hc            & 0.0\%  \\
        Discharge to SNF-Medicaid Only Certified    & 0.0\%  \\
    \bottomrule \end{tabular}
    \label{tab:acuity_stats_stats}
\end{table}

\paragraph{Next Timepoint: WBM} \hfill
\\ \textbf{Description:} We predict two local autoregressive tasks designed to assess the model's ability to forecast what will happen in the immediate next hour. First, we predcit which labs \& vitals will be measured in the next hour via multi-label binary classification; second, we predict what values will be observed for those labs \& vitals that are measured via continuous regression. For reporting purposes, we present only the classification task, as our regression performance was universally poor, save on a subset of commonly measured labs/vitals (in particular, blood pressures, oxygen saturation, and heart rate). However, it is included in training ensembles as, in prior experiments, we observed that, surprisingly, removing it actually weakened the overall ensemble.
\\ \textbf{Data Source:} This is sourced directly from the time-varying hourly input features output by MIMIC Extract~\cite{wang_mimic-extract:_2019}.
\\ \textbf{Prior Art:} Predicting which labels will be measured in the next time-point has been explored using reinforcement learning  \cite{chang19a}.
\\ \textbf{Statistics:} The labs \& vitals over which we predict, along with their observed measurement rates and average continuous values are shown in Table~\ref{tab:next_timepoint_stats}.

\begin{table}
    \centering \small
    \caption{The Labs \& Vitals we predict over for our next timepoint task, along with the \% of time they are measured. These are also the input labs \& vitals we use as the input to our pipeline. Note that all inputs were centered and scaled to unit variance when measured, so our continuous regression task had ouputs that were 0 mean and unit variance.}
    \begin{tabular}{lr} \toprule
        Lab / Vital                               & Measurement Rate (\%) \\ \midrule
        Heart Rate                                &              91.6\% \\
        Respiratory Rate                          &              90.2\% \\
        Diastolic Blood Pressure                  &              88.8\% \\
        Systolic Blood Pressure                   &              88.8\% \\
        Mean Blood Pressure                       &              88.3\% \\
        Oxygen Saturation                         &              87.6\% \\
        Temperature                               &              29.8\% \\
        Glucose                                   &              23.2\% \\
        Central Venous Pressure                   &              20.3\% \\
        Glascow Coma Scale Total                  &              17.7\% \\
        Hematocrit                                &              11.2\% \\
        Potassium                                 &              10.4\% \\
        Sodium                                    &               9.9\% \\
        Pulmonary Artery Pressure Systolic        &               9.4\% \\
        Chloride                                  &               9.4\% \\
        Ph                                        &               9.2\% \\
        Hemoglobin                                &               9.0\% \\
        Creatinine                                &               8.8\% \\
        Blood Urea Nitrogen                       &               8.7\% \\
        Bicarbonate                               &               8.6\% \\
        Magnesium                                 &               8.3\% \\
        Anion Gap                                 &               8.3\% \\
        Partial Pressure Of Carbon Dioxide        &               8.3\% \\
        Co2 (Etco2, Pco2, Etc.)                   &               8.3\% \\
        Platelets                                 &               8.2\% \\
        Positive End-Expiratory Pressure Set      &               8.0\% \\
        White Blood Cell Count                    &               7.9\% \\
        Calcium                                   &               7.1\% \\
        Fraction Inspired Oxygen Set              &               7.0\% \\
        Tidal Volume Observed                     &               6.8\% \\
        Mean Corpuscular Hemoglobin Concentration &               6.2\% \\
        Mean Corpuscular Volume                   &               6.2\% \\
        Red Blood Cell Count                      &               6.2\% \\
        Mean Corpuscular Hemoglobin               &               6.2\% \\
        Partial Thromboplastin Time               &               6.0\% \\
        Prothrombin Time Inr                      &               5.7\% \\
        Prothrombin Time Pt                       &               5.7\% \\
        Peak Inspiratory Pressure                 &               5.6\% \\
        Phosphate                                 &               5.5\% \\
        Phosphorous                               &               5.4\% \\
        Respiratory Rate Set                      &               4.9\% \\
        Calcium Ionized                           &               4.9\% \\
        Fraction Inspired Oxygen                  &               4.7\% \\
        Tidal Volume Set                          &               4.6\% \\
        Partial Pressure Of Oxygen                &               4.3\% \\
        Cardiac Index                             &               3.6\% \\
        Co2                                       &               3.5\% \\
        Pulmonary Artery Pressure Mean            &               3.5\% \\
        Tidal Volume Spontaneous                  &               3.5\% \\
        Plateau Pressure                          &               3.4\% \\
        Systemic Vascular Resistance              &               3.4\% \\
        Potassium Serum                           &               3.2\% \\
        Cardiac Output Thermodilution             &               3.0\% \\
        Lactate                                   &               2.7\% \\
        Weight                                    &               2.5\% \\
        Lactic Acid                               &               2.4\% \\
    \bottomrule \end{tabular}
    \label{tab:next_timepoint_stats}
\end{table}

\begin{figure}
    \centering
    \includegraphics[width=0.6\linewidth]{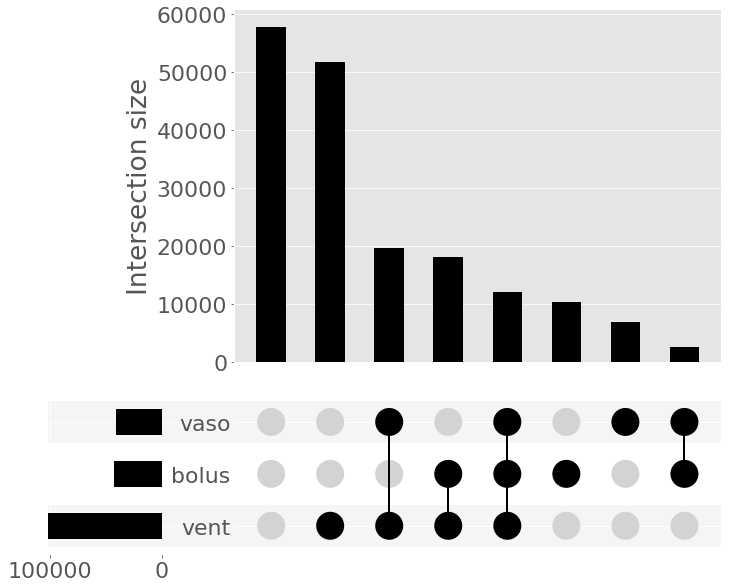}
    \caption{An Upset plot showing the frequency of relative combinations of our three treatment types: Vasopressors (vaso), Ventilation (vent), and Fluid Bolus administration (bolus).}
    \label{fig:treatment_combos}
\end{figure}

\paragraph{Future Treatment Sequence: FTS} \hfill
\\ \textbf{Description:} This task is a global, autoregressive task designed to force the model to learn how to predict the high-level future of the patient's care. Labels for this task are the sequence of treatment combinations the patient will receive over the remainder of their stay, in a duration agnostic manner. In our context, individual treatments are aggregated first into the broad categories \emph{ventilation}, \emph{vasopressors}, and \emph{fluid boli}, so each sequential label is an element of the powerset of these categories. This sequence of treatment sets is \emph{duration agnostic}---i.e., elements in the label sequence merely represent that a patient will receive this combination of treatments next in sequence, and do not comment on for how long the patient will receive them. Accordingly, there are no sequential duplicates in this label sequence.
The task-specific decoder is an LSTM recurrent neural network, and during both training and evaluation we use teacher forcing~\cite{lamb2016professor}---i.e., we pass in the true sequence of treatments when asking the system to predict the next element. The input from the encoder is used as the initial hidden state to this decoder model.  This means that our evaluation results \emph{should not} be interpreted as the model's ability to correctly decode the full projection of future treatments, but rather the model's ability to understand how clinicians will transition between treatments given this patient's unique record to date. Changing the formulation of this task to not use teacher forcing during evaluation would make the task much more difficult, and represents a valuable area of future work. This task is, to the best of our knowledge, novel.
\\ \textbf{Data Source:} We construct labels for this task ourselves based on the provided time-varying treatments produced by MIMIC Extract.
\\ \textbf{Prior Art:} While this task directly has not been explored previously, various researchers have investigated learning optimal control policies for applications of treatments, including ventilators or vasopressors \cite{Wu_vassopressur_2016, peng2018improving,10.1093/jamia/ocu051}.
%\todo{check citations, cite RL works if not already.}
\\ \textbf{Statistics:} We show the relative frequency of the various treatment combinations (recall that our labels here are \emph{subsets} of treatments, expanded into a one-hot encoding over the entire powerset) in Figure~\ref{fig:treatment_combos}.

\section{Hyperparameter Search Analysis}
\label{sec:hyperparameter_details}
\subsection{Hyperparameter Search Algorithm Details}
We used the bayesian hyperparameter tuning \texttt{Hyperopt} library~\cite{bergstra_making_2013}---specifically, \texttt{Hyperopt}s Tree of Parzen Estimators (TPE) optimization method. All hyperparameter tuning was performed to optimize for the average AUROC (or, for our regression task, an AUROC analog score defined to be $2^{R^2 - 1}$) across all tasks and labels under full MT training. We allowed \texttt{Hyperopt} to tune both the hyperparameters for all architectures as well as select which architecture (GRU, Linear baseline, Transformer) should be used. The system was permitted to devote more samples to higher performing architectures. In order to ensure that all architectures were sampled to a reasonable degree, we did a separate, albeit smaller run of \texttt{Hyperopt} iterations on each architecture independently, allowing the system to use the prior samples of that architecture from the joint training to inform its algorithm's next hyperparameter selection. We also performed a mild amount of manual hyperparameter selection early in the process, largely to adjust sampling distributions for the \texttt{Hyperopt} search space before the final overall hyperparameter tuning run.

For final runs, we used the best performing (as evaluated on our 10\% validation set) hyperparameters found across all iterations for each architecture independently. Final selections are discussed below.

\subsection{Expanded Hyperparameter Search Biases Discussion}
The training procedure described above was chosen for computational efficiency; by jointly optimizing over all tasks and architectures in a systematic, motivated fashion we can ensure we have good coverage over all likely high-performing parameters for all tasks and architectures, while dramatically minimizing our overall search time. However, this does induce two potential biases. 

First, not all architectures received the same number of samples. In particular, architectures that both (1) took longer to run, and (2) seemed less promising in early experiments would receive fewer tuning samples. This may disproportionately penalize, for example, the transformer model, which both took the longest to run and had poor preliminary results.

Second, this system may favor multi-task based systems over single-task systems, potentially in a task specific manner. This is a particularly poignant concern, given that MTL has been postulated to have a regularizing effect in past literature~\cite{ruder2017overview}; thus, we might be concerned in this situation that the multi-task chosen hyperparameters would inappropriately prefer less regularization than a true optimal ST model would. We note two mitigating factors that make us \emph{not} concerned that this bias plays a serious role. First, our ST models \emph{already} outperform the full MT model; thus, our results do not appear concordant with this bias. Second, we additionally performed a (admittedly smaller) secondary round of hyperparameter tuning on a single task alone, and analyzed the results of our full hyperparamaeter tuning system as though we were optimizing for that same task, and in both cases found the performance difference between the chosen optimal parameters negligible.

\subsection{Search Space}
For our hyperparameter search procedure, we searched over a wide variety of parameters, including number of epochs, batch size, learning rate, learning rate decay paradigms, L2 regularization penalty, dropout, a weighting for the regression task losses' contribution to training, the maximum length of a patients record included, the size, number, and configuration of various hidden layers, pooling and fully connected stack parameters, and various other model-specific options. All search distributions are shown in Table~\ref{tab:hyperparameter_search_space}. Note that some of these search space parameters are specific to our implementation; for example, the ``Encoder Hidden Size Multiplier'' used in the transformer architecture encodes the relationship between the size of the overall internal transformer hidden state and the number of attention heads (the former must be divisible by the latter). Additionally, note that this search distribution reflects our \emph{final} search distribution, used to enable the most granular refinement, while optimal parameters may have been chosen for certain model types by earlier incarnations of the search on wider, more uncertain search distributions, or search distributions with fewer options enabled.
% For example, the regression task weight of our optimal GRU model is \emph{precisely} 1; not because this was found with the distribution below, but rather because our prior best validation-set parameters outperformed all samples in the new, refined distribution (albeit often only slightly) and those parameters were searched without the ability to tune the regression weight.

\begin{table}
    \centering
    \caption{
    The \texttt{Hyperopt} search space we used in this work. Distributions are noted in pseudocode, but typically refer directly to the appropriate analog in \texttt{Hyperopt} (e.g., a uniform distribution over an integral parameter maps to the quantized uniform distribution that only outputs integers). \emph{Shared} hyperparameters were used across all 3 model types. The linear model required no non-shared hyperparameters.
    }
    \begin{tabular}{llr} \toprule
        Architecture                    & Hyperparameter                    & Search Space \\ \midrule
        \multirow[l]{9}{*}{Shared}      & \# Epochs                         & \texttt{Uniform[15, 30]} \\
                                        & Batch Size                        & \texttt{Uniform[4, 64]} \\
                                        & Learning Rate (LR)                & \texttt{Lognormal[-7, 0.5]} \\
                                        & LR Decay                          & \texttt{Loguniform[-2.3, 0]} \\
                                        & LR Step                           & \texttt{Uniform[1, 25]} \\
                                        & Hidden Dropout                    & \texttt{Uniform[0, 0.5]} \\
                                        & Hidden Size                       & \texttt{Uniform[8, 256]} \\
                                        & Weight Decay                      & \texttt{Uniform[0, 1]} \\
                                        % & Regression Task Weight            & \texttt{Uniform[0, 1]} \\
                                        & Input Window Size (h)             & \texttt{Uniform[12, 168]} \\ \midrule
        \multirow[l]{5}{*}{Transformer} & Encoder Hidden Size Multiplier    & \texttt{Uniform[4, 32]} \\
                                        & Intermediate Size                 & \texttt{Uniform[32, 256]} \\
                                        & \# Attention Heads                & \texttt{Uniform[2, 24]} \\
                                        & \# Hidden Layers                  & \texttt{Uniform[1, 4]} \\
                                        & Use CLS Analog                    & \texttt{Choice[True, False]} \\ \midrule
        \multirow[l]{7}{*}{GRU}         & Bidirectional                     & \texttt{Choice[True, False]} \\
                                        & \# Hidden Layers                  & \texttt{Uniform[1, 3]} \\
                                        & Encoder Hidden Layer Size         & \texttt{Uniform[16, 512]} \\
                                        & Encoder \# Fully Connected Layers & \texttt{Uniform[0, 3]} \\
                                        & GRU Pooling Method                & \texttt{Choice[max, avg, last]} \\
                                        & GRU FC Layer Base Size            & \texttt{Uniform[32, 512]} \\
                                        & GRU FC Layer Growth               & \texttt{Loguniform[-1.1, 1.1]} \\
    \bottomrule \end{tabular}
    \label{tab:hyperparameter_search_space}
\end{table}

\subsection{Final Optimal Parameters}
\paragraph{Projection Model}
Our projected model ran for 22 epochs, using a projection dimension of 140, a batch size of 16 and learning rate of 0.00024 with no decay, and dropout at 0.22.
% {"max_seq_len": 48, "modeltype": "linear", "epochs": 22, "batches_per_gradient": 1  "in_dim": 32, "hidden_size": 140, "intermediate_size": 128,  "batch_size": 16, "learning_rate": 0.00024349156699823263, "learning_rate_decay": 1, "learning_rate_step": 1,  "hidden_dropout_prob": 0.2239657179320786, 
% }

\paragraph{GRU}
Our optimal GRU model was a unidirectional, 2 126-dimensional hidden layer GRU which ran for 18 epochs with an effective batch size of 254, a projection dimension of 233, a learning rate of 0.001 with no decay and dropout of 0.42. It used the last-element GRU pooling, and had no fully connected layers post-processing the GRU samples.
% {"max_seq_len": 48, "modeltype": "GRU", "epochs": 18, "batches_per_gradient": 7, "in_dim": 32,
%  "hidden_size": 233, "intermediate_size": 128, "num_attention_heads": 4, "num_hidden_layers": 2, "batch_size": 32, "learning_rate": 0.001343795748862106,  "gru_num_hidden": 2, "gru_hidden_layer_size": 126,  "hidden_dropout_prob": 0.41699363848857535, 
%  "pooling_method": "max", "pooling_kernel_size": 4, "pooling_stride": null,  "do_bidirectional": false}
 
\paragraph{Transformer}
Our optimal transformer ran for 24 epochs with a projection dimensionality of 72, 12 attention heads, 1 hidden layer, an intermediate size of 55, a specially added overall sequence sentinel token, a learning rate of 0.002, a batch size of 30, and dropout of 0.18.

\section{Full Architecture Details for Linear Baseline \& Transformer Architectures}
\paragraph{Linear Baseline Projection Architecture} For our linear baseline projection architecture, we simply project all inputs into the same embedding space, concatenate the full input sequence together (as we used a fixed input window size, this is a fixed-size representation), then pass that through to the per-task decoders. Note that the only ``multi-tasking'' that happens in this representation is that the projection layers mapping content to the input embedding spacea are shared.

\paragraph{Transformer} For our transformer architecture, we used a bidirectional transformer (e.g., a BERT architecture~\cite{devlin_bert:_2018}) operating on the continuous (projected) embeddings of all input features. We did not employ any positional embeddings; however, we did add an auxiliary ``CLS'' token to the front of each sequence which was used as the source of our pooled representation (much like BERT). This strategy of pooling was found to be preferred in our hyperparameter tuning.

\section{GRU Population Imbalance Experiments}
\label{sec:fairness}
We examined our GRU model under population imbalance in the following manner. First, we pre-trained a multi-task model on full, unaltered data. Next, we fine-tuned the model on a dataset that was randomly subsampled in a genotypical sex imbalanced manner, up to and including removing all patients who were genotypically female. We anticipated that for single-task models (which were not pre-trained, and instead tuned from scratch on these imbalanced datasets), the use of imbalanced data would engender significant biases in model performance favoring the majority class (genotypically male patients) and that the balanced multi-task pre-training would help ameliorate these biases. However, using AUROC discrepancy between genotypically male and female patients as our guide, we did not see either effect. In Table~\ref{tab:gender_bias_results}, we show the Male - Female AUROC discrepancy of our GRU model under the ST, FTD, and FTF training regimes in the case that all genotypically female patients were removed during fine-tuning. We see that in roughly half the case, this discrepancy is positive, and roughly half it is negative (indicating the model about equally favors genotypically male patients and genotypically female patients), and that the discrepancies are largely small. Unfortunately, due to time constraints, we were only able to run one sample of these runs, so it may be that a stronger effect would emerge with more samples and greater statistical power; however, these preliminary findings do not suggest that this is the case.

\begin{table}
    \centering
    \caption{Comparison of the $\textrm{Male} - \textrm{Female}$ AUROC for fine-tuned models on data biased to have no women in the training set. Bold indicates lowest discrepancy (e.g., most favorable to women).}
    \begin{tabular}{lrrr}
    \toprule
    {} & \multicolumn{3}{l}{GRU} \\
    {} &                   ST &                  FTD &                  FTF \\
    Task &                      &                      &                      \\
    \midrule
    MOR  &                $0.5$ &               $-0.2$ &  $\boldsymbol{-1.1}$ \\
    CMO  &                $3.1$ &               $-0.1$ &  $\boldsymbol{-2.8}$ \\
    DNR  &                $0.1$ &                $3.9$ &  $\boldsymbol{-1.0}$ \\
    DIS  &  $\boldsymbol{-0.3}$ &                $0.6$ &                $2.8$ \\
    ICD  &   $\boldsymbol{1.9}$ &                $4.9$ &                $3.7$ \\
    LOS  &               $-2.9$ &  $\boldsymbol{-3.5}$ &                $1.1$ \\
    REA  &  $\boldsymbol{-3.4}$ &                $7.9$ &               $-0.2$ \\
    ACU  &               $-0.7$ &               $-0.0$ &  $\boldsymbol{-3.8}$ \\
    WBM  &  $\boldsymbol{-0.2}$ &                $0.4$ &                $0.6$ \\
    FTS  &  $\boldsymbol{-2.0}$ &               $-1.7$ &               $-1.2$ \\
    \bottomrule
    \end{tabular}    \label{tab:gender_bias_results}
\end{table}

\section{Few-Shot Experiments under all architectures}
\label{sec:few_shot}

\begin{figure}
    \centering
    \includegraphics[width=\linewidth]{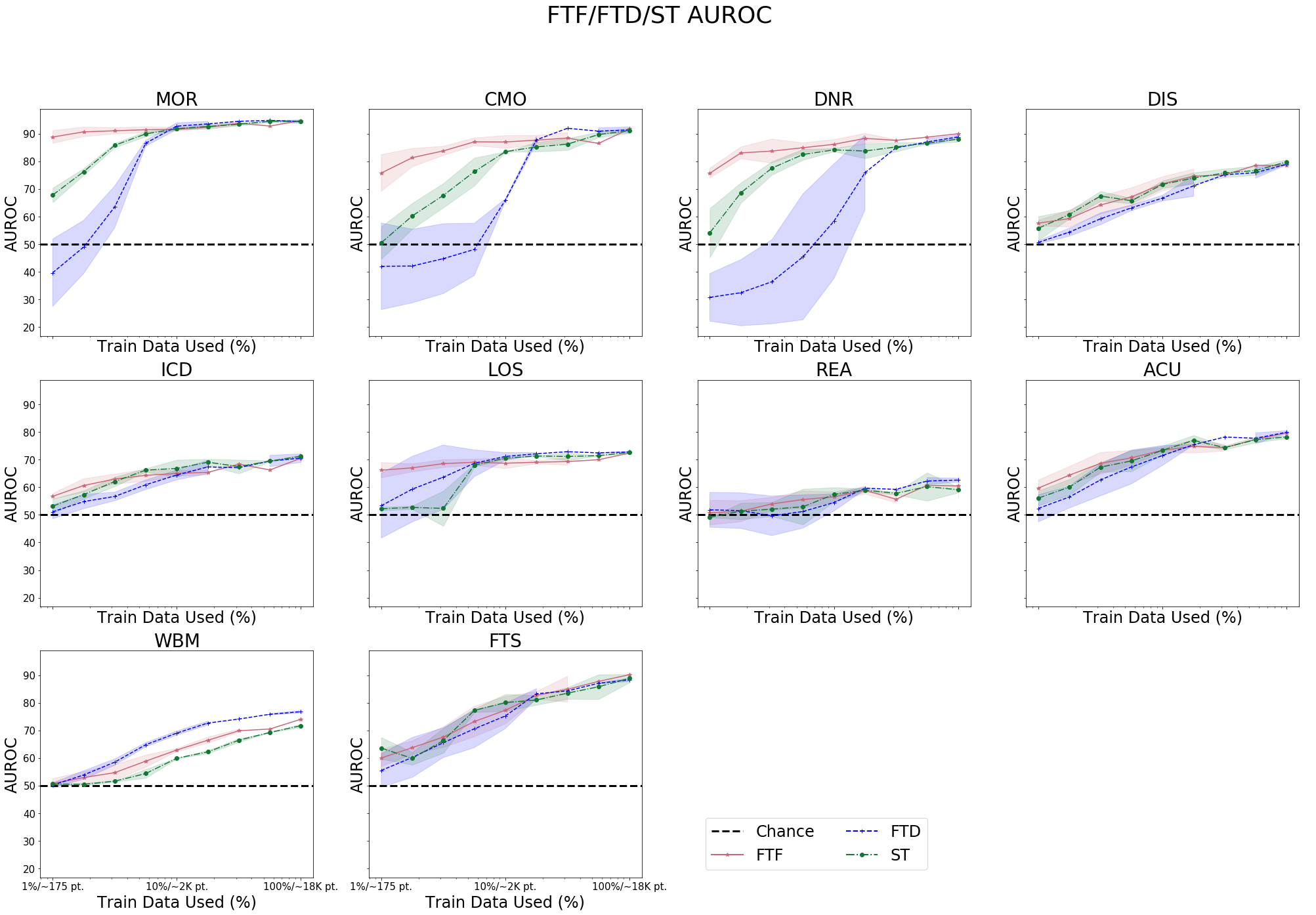}
    \caption{Few-shot experiments for GRU (duplicated from main body)}
    \label{fig:small_data_gru}
\end{figure}

\begin{figure}
    \centering
    \includegraphics[width=\linewidth]{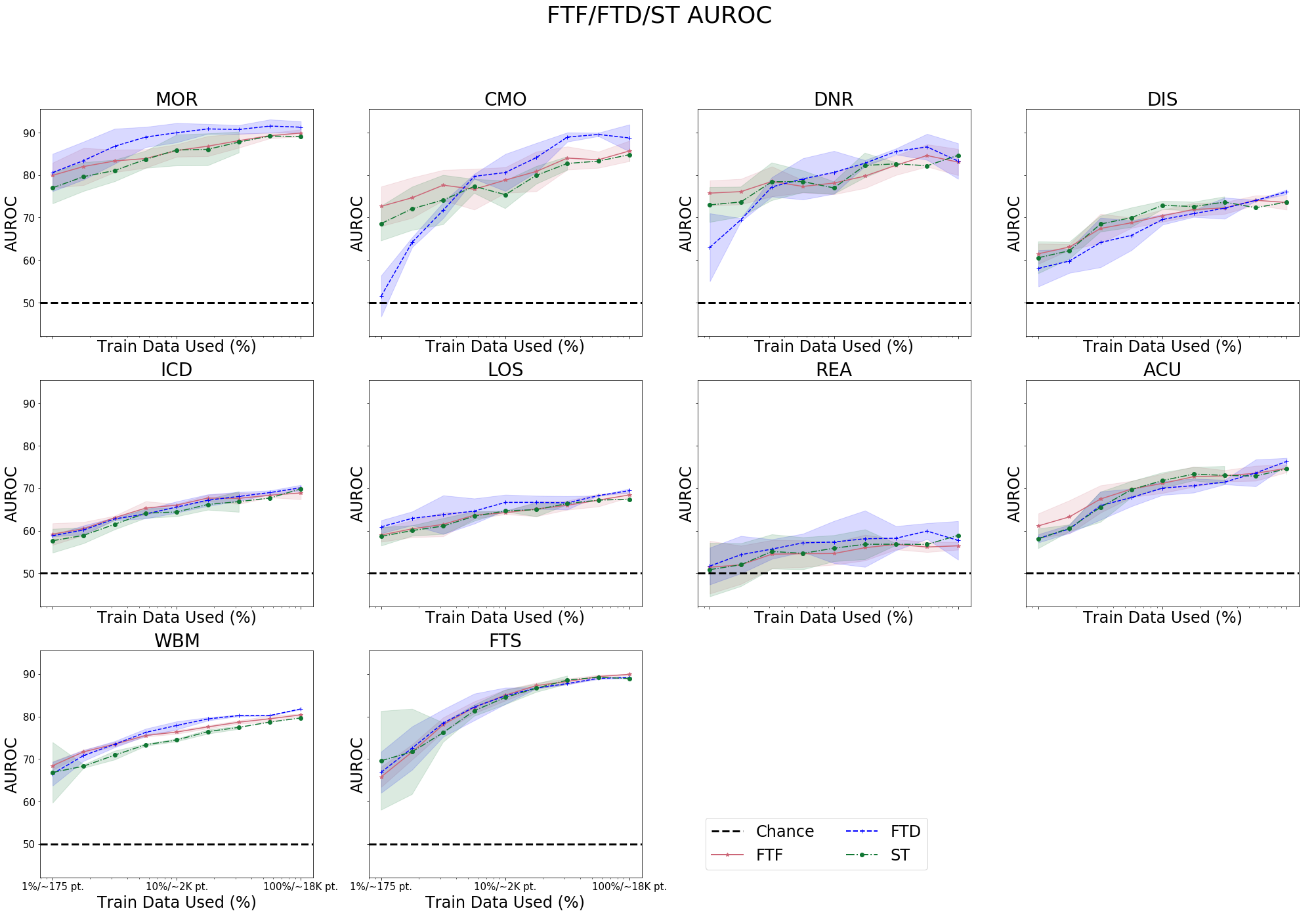}
    \caption{Few-shot experiments for the linear projection model}
    \label{fig:small_data_proj}
\end{figure}

\begin{figure}
    \centering
    \includegraphics[width=\linewidth]{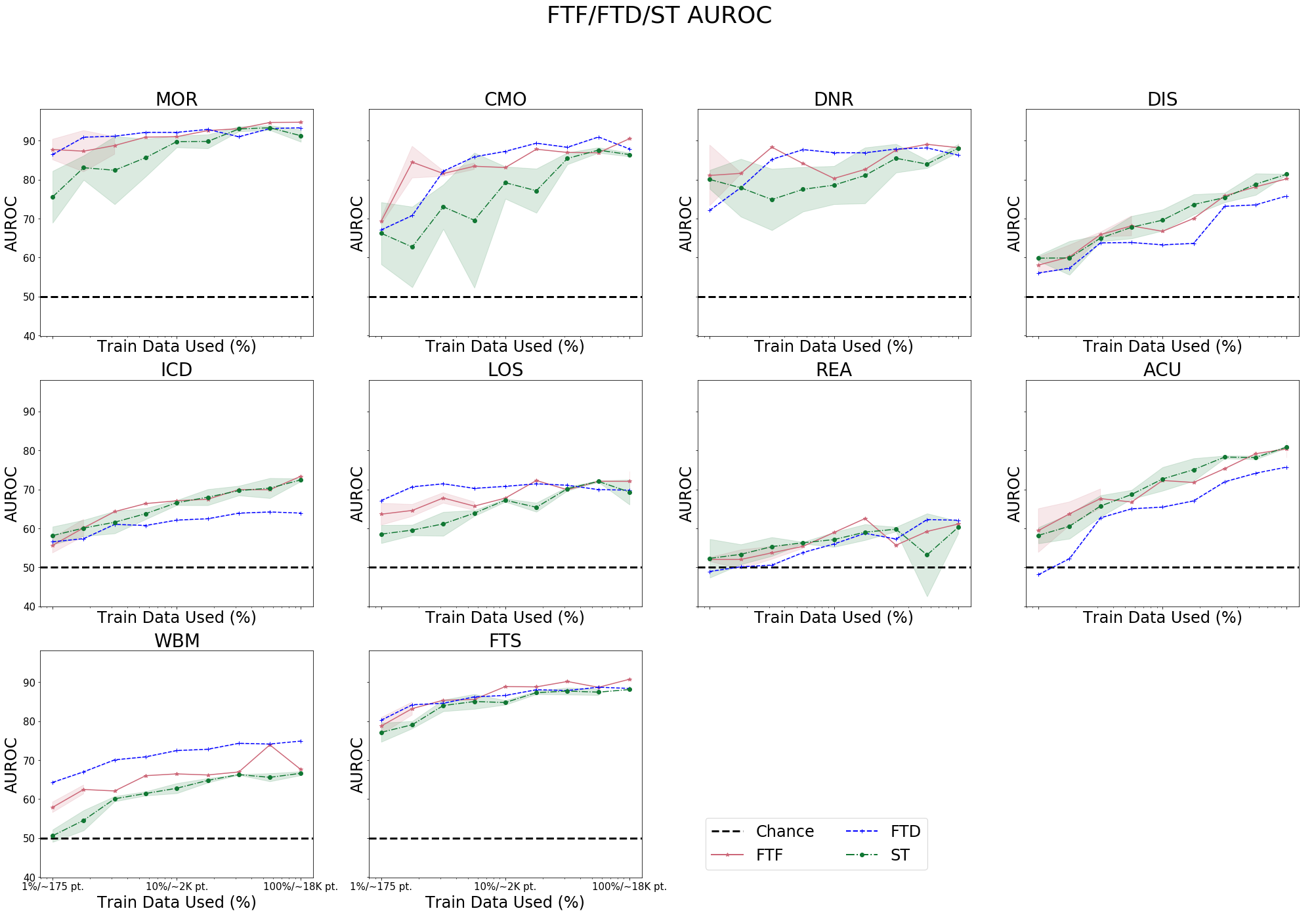}
    \caption{Few-shot experiments for the tranformer model}
    \label{fig:small_data_trans}
\end{figure}

We replicated our few-shot experiments under all 3 architectures, and present here in Figures~\ref{fig:small_data_gru}, \ref{fig:small_data_proj}, \ref{fig:small_data_trans} plots showing the performance of ST, FTF, and FTD model types on subsampled datasets ranging from 1\% to 100\% across each task individually. We can see that while the dramatic gains of the GRU model are not replicated on other model types, on the transformer model, we see consistent gains of fine-tuning regimes on MOR, CMO, LOS, and WBM, much like the GRU. Additionally, and somewhat surprisingly, while the GRU outperformed the transformer architecture on the full dataset as a general rule, in the very small data regime for some tasks, the transformer offers notable gains (e.g., MOR). The linear projection model shows largely concordant behavior across all model types, with perhaps a slight gain on a subset of tasks, including MOR and WBM, towards fine-tuning results.


\section{Full Results}
\label{sec:all_arch_results}
\begin{figure}
    \centering
    \includegraphics[width=\textwidth]{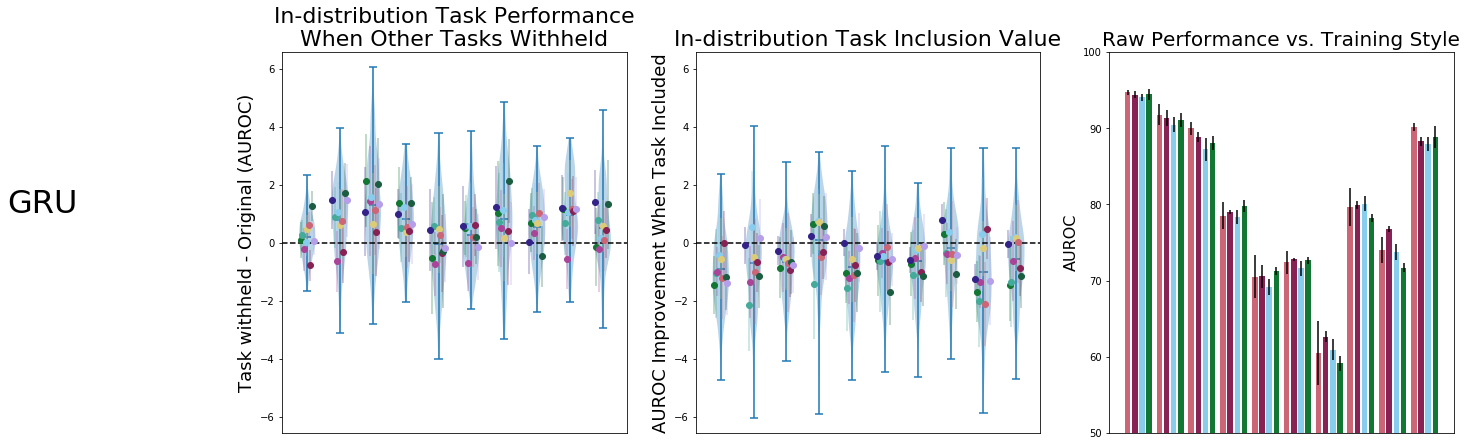}
    
    \includegraphics[width=\textwidth]{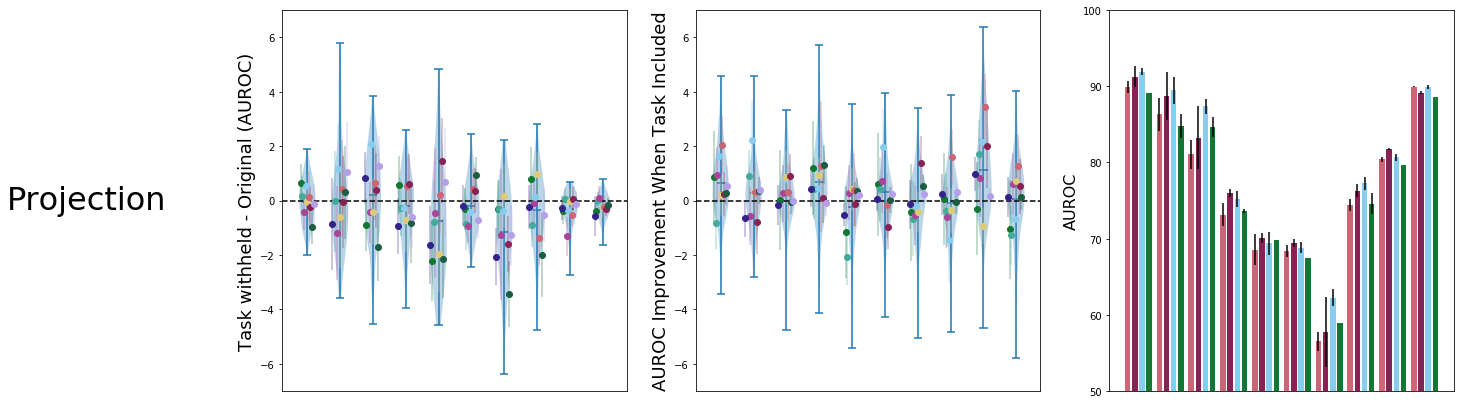}
    
    \includegraphics[width=\textwidth]{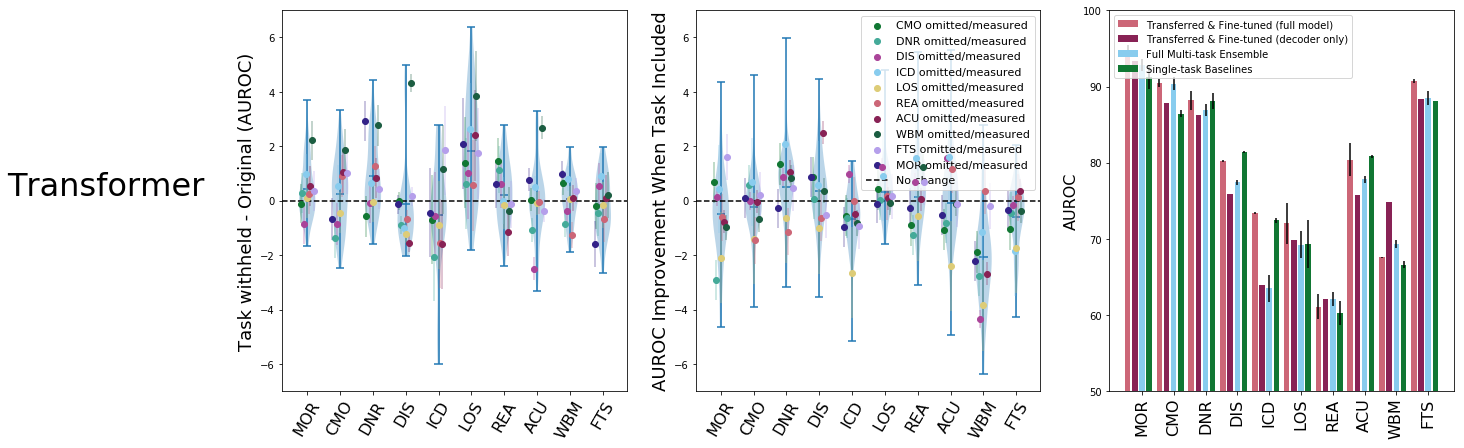}
    \caption{\emph{Left}: The performance delta (positive meaning better than fully multi-task performance) on each of our tasks ($x$-axis) when the various \emph{other} tasks are held out (point color; horizontal offset of colored points within each task column is only so the task-specific error bars can be simultaneously visible, and is consistent, but arbitrary), all measured in AUROC \%. This shows that task performance commonly improves when the other tasks are held out. \emph{Middle}: The average improvement across \emph{all} other tasks (point color; horizontal offset within each task is merely for display purposes) when each of the possible held-out task groups ($x$-axis) are included in the ensemble (positive meaning inclusion of this task helps). We see that omitting single task often improves the performance on other tasks. \emph{Right}: Fully multi-task performance vs. single task performance vs. fine-tuned performance. Fine-tuned or multi-task representations are quite consistently preferred. \emph{Top:} The GRU model architecture. \emph{Middle:} The projection model architecture. \emph{Bottom:} The Transformer model architecture.}
    \label{fig:all_arch_results}
\end{figure}

Figure~\ref{fig:all_arch_results} shows full results for all models in the full data regime, including local negative transfer analyses (analogous to Section 5.1 in the main body) in the two left-most columns, and global results in the right-most column. We can see several main take-aways from these results. First, while the transformer and GRU results both display significant common negative transfer, as evidenced by the majority of the violin-plot point mass being above 0 in the left-most column, the linear projection algorithm actually displays different behavior, with a significant extent of positive transfer happening as well. This is also reflected in the overall results globally --- for the linear system, unlike the GRU and Transformer results, the full multi-task system is actually commonly one of the top performing system, and outperforms the ST model on all tasks except ICD code prediction.

Secondly, we can see that MTL preferences are very architecture dependent, with common discrepancies in which model types are preferred across the different architectures. This suggests researchers should be cognizant of strong relationships between MTL efficacy and architecture choice in future modelling. Lastly, we see that there is significant, task-dependent variability in the Transformer results, much moreso than the other model types. For example, we can see that including the WBM task is extremely harmful for the DIS and LOS tasks, inducing a roughly 4\% drop in AUROC in each, as well as inducing smaller costs to other tasks. This relationship is nowhere near as pronounced for the GRU model, and is almost reversed for the linear model. This may be related to the fact that, while the GRU model is our best performing model on average, the linear model outperforms both other model types on the WBM task by significant margins.

\paragraph{Number of independent repeats} To assess variance in all our results, we ran a number of independent repeats of each run under different random seeds. Note that we did not alter the train/val/test split for these runs, as differing splits would require re-doing hyperparameter tuning which was computationally prohibitive. Total numbers of samples for all runs is shown in Tables~\ref{tab:num_samples_gru}, \ref{tab:num_samples_linear}, and \ref{tab:num_samples_self_attention}. Additional samples, both repeats and train/val/test splits / hyperparameter tuning runs would be run prior to camera ready submission.

\setlength\tabcolsep{1.5pt} % default value: 6pt

\begin{sidewaystable}
    \tiny
    \centering
    \caption{Number of samples run for all modes of the GRU architecture. Additional samples will be run before any camera ready publication.}
\begin{tabular}{lrrrrrrrrrrrrrrrrrrrrrrrrrrrrrrrrrrrrrrrrrrr}
\toprule
{} &        MT &        ST &       FTD &       FTF & \multicolumn{13}{l}{ST Few-Shot \%} & \multicolumn{13}{l}{FTD Few-Shot \%} & \multicolumn{13}{l}{FTF Few-Shot \%} \\
{} & \multicolumn{4}{l}{Full-data} &           0.0 & 0.1 & 0.2 & 0.3 & 0.6 & 1.0 & 1.8 & 3.2 & 5.6 & 10.0 & 17.8 & 31.6 & 56.2 &            0.0 & 0.1 & 0.2 & 0.3 & 0.6 & 1.0 & 1.8 & 3.2 & 5.6 & 10.0 & 17.8 & 31.6 & 56.2 &            0.0 & 0.1 & 0.2 & 0.3 & 0.6 & 1.0 & 1.8 & 3.2 & 5.6 & 10.0 & 17.8 & 31.6 & 56.2 \\
\midrule
MOR &        10 &         3 &         3 &         5 &             5 &   5 &   5 &   5 &   4 &   2 &   2 &   2 &   2 &    2 &    2 &    2 &    2 &              2 &   3 &   3 &   3 &   2 &   3 &   3 &   2 &   2 &    2 &    2 &    1 &    2 &              4 &   5 &   5 &   5 &   4 &   5 &   5 &   5 &   5 &    5 &    4 &    3 &    1 \\
CMO &        10 &         3 &         3 &         5 &             5 &   5 &   5 &   5 &   4 &   3 &   3 &   3 &   3 &    3 &    3 &    3 &    3 &              2 &   3 &   3 &   3 &   2 &   3 &   3 &   3 &   2 &    2 &    2 &    1 &    2 &              4 &   5 &   5 &   5 &   4 &   5 &   5 &   5 &   5 &    5 &    3 &    4 &    1 \\
DNR &        10 &         3 &         3 &         5 &             5 &   5 &   5 &   5 &   4 &   3 &   3 &   3 &   3 &    3 &    2 &    2 &    2 &              2 &   3 &   3 &   3 &   2 &   3 &   3 &   3 &   3 &    3 &    2 &    1 &    2 &              4 &   5 &   5 &   5 &   4 &   5 &   5 &   5 &   5 &    5 &    3 &    3 &    1 \\
ICD &        10 &         3 &         3 &         5 &             5 &   5 &   5 &   5 &   4 &   2 &   2 &   2 &   2 &    2 &    2 &    2 &    2 &              2 &   3 &   3 &   3 &   2 &   3 &   3 &   2 &   2 &    2 &    2 &    1 &    2 &              4 &   5 &   5 &   5 &   4 &   5 &   5 &   5 &   5 &    5 &    3 &    1 &    1 \\
LOS &        10 &         3 &         3 &         5 &             5 &   5 &   5 &   5 &   4 &   2 &   2 &   2 &   2 &    2 &    2 &    2 &    2 &              2 &   3 &   3 &   3 &   2 &   3 &   3 &   2 &   2 &    2 &    2 &    1 &    2 &              4 &   5 &   5 &   5 &   4 &   5 &   5 &   5 &   5 &    5 &    4 &    3 &    1 \\
REA &        10 &         3 &         3 &         5 &             5 &   5 &   5 &   5 &   4 &   2 &   2 &   2 &   2 &    2 &    2 &    2 &    2 &              2 &   3 &   3 &   3 &   2 &   3 &   3 &   2 &   2 &    2 &    2 &    1 &    2 &              4 &   5 &   5 &   5 &   4 &   5 &   5 &   5 &   5 &    5 &    4 &    3 &    1 \\
DIS &        10 &         3 &         3 &         5 &             5 &   5 &   5 &   5 &   4 &   3 &   3 &   3 &   3 &    3 &    3 &    2 &    2 &              2 &   3 &   3 &   3 &   2 &   3 &   3 &   3 &   2 &    2 &    3 &    1 &    2 &              4 &   5 &   5 &   5 &   4 &   5 &   5 &   5 &   5 &    5 &    2 &    1 &    1 \\
ACU &        10 &         3 &         3 &         5 &             5 &   5 &   5 &   5 &   4 &   3 &   3 &   3 &   3 &    3 &    3 &    2 &    2 &              2 &   3 &   3 &   3 &   2 &   3 &   3 &   3 &   2 &    2 &    2 &    1 &    2 &              4 &   5 &   5 &   5 &   4 &   5 &   5 &   5 &   5 &    5 &    4 &    3 &    1 \\
WBM &        10 &         3 &         3 &         5 &             5 &   5 &   5 &   5 &   4 &   2 &   2 &   2 &   2 &    2 &    2 &    2 &    2 &              2 &   3 &   3 &   3 &   2 &   3 &   3 &   3 &   3 &    3 &    2 &    1 &    2 &              4 &   5 &   5 &   5 &   4 &   5 &   5 &   5 &   5 &    5 &    3 &    3 &    1 \\
FTS &        10 &         3 &         3 &         5 &             5 &   5 &   5 &   5 &   4 &   3 &   3 &   3 &   2 &    2 &    2 &    2 &    2 &              2 &   3 &   3 &   3 &   2 &   3 &   3 &   3 &   2 &    2 &    2 &    1 &    2 &              4 &   5 &   5 &   5 &   4 &   5 &   5 &   5 &   5 &    5 &    4 &    3 &    1 \\
\bottomrule
\end{tabular}
    \label{tab:num_samples_gru}
\end{sidewaystable}

\begin{table}
    \tiny
    \centering
    \caption{Number of samples run for all modes of the linear architecture. Additional samples will be run before any camera ready publication.}
\begin{tabular}{lrrrrrrrrrrrrrrrrrrrrrrrrrrrrrrrrrrrrr}
\toprule
{} &        MT &        ST &       FTD &       FTF & \multicolumn{11}{l}{ST Few-Shot \%} & \multicolumn{11}{l}{FTD Few-Shot \%} & \multicolumn{11}{l}{FTF Few-Shot \%} \\
{} & \multicolumn{4}{l}{Full-data} &           0.2 & 0.3 & 0.6 & 1.0 & 1.8 & 3.2 & 5.6 & 10.0 & 17.8 & 31.6 & 56.2 &            0.2 & 0.3 & 0.6 & 1.0 & 1.8 & 3.2 & 5.6 & 10.0 & 17.8 & 31.6 & 56.2 &            0.2 & 0.3 & 0.6 & 1.0 & 1.8 & 3.2 & 5.6 & 10.0 & 17.8 & 31.6 & 56.2 \\
\midrule
MOR &         9 &         1 &         2 &         5 &             4 &   4 &   3 &   5 &   5 &   5 &   5 &    2 &    3 &    3 &    1 &              2 &   2 &   2 &   2 &   2 &   2 &   2 &    2 &    2 &    2 &    2 &              5 &   5 &   4 &   5 &   5 &   5 &   5 &    4 &    5 &    5 &    5 \\
CMO &         9 &         2 &         2 &         5 &             3 &   3 &   2 &   5 &   5 &   5 &   5 &    4 &    4 &    2 &    1 &              2 &   2 &   2 &   2 &   2 &   2 &   2 &    2 &    2 &    2 &    2 &              5 &   5 &   4 &   5 &   5 &   5 &   5 &    4 &    5 &    4 &    4 \\
DNR &         9 &         2 &         2 &         5 &             4 &   3 &   2 &   5 &   5 &   5 &   5 &    4 &    3 &    2 &    1 &              2 &   2 &   2 &   2 &   2 &   2 &   2 &    2 &    2 &    2 &    2 &              5 &   5 &   4 &   5 &   5 &   5 &   5 &    4 &    5 &    4 &    4 \\
ICD &         9 &         1 &         2 &         5 &             4 &   4 &   3 &   5 &   5 &   5 &   4 &    2 &    3 &    3 &    1 &              2 &   2 &   2 &   2 &   2 &   2 &   2 &    2 &    2 &    2 &    2 &              5 &   5 &   4 &   5 &   5 &   5 &   5 &    4 &    5 &    5 &    5 \\
LOS &         9 &         1 &         2 &         5 &             4 &   4 &   3 &   4 &   4 &   4 &   4 &    3 &    4 &    3 &    1 &              2 &   2 &   2 &   2 &   2 &   2 &   2 &    2 &    2 &    2 &    2 &              5 &   5 &   4 &   5 &   5 &   5 &   5 &    4 &    5 &    5 &    4 \\
REA &         9 &         1 &         2 &         5 &             4 &   4 &   2 &   5 &   4 &   4 &   4 &    3 &    3 &    2 &    1 &              2 &   2 &   2 &   2 &   2 &   2 &   2 &    2 &    2 &    2 &    2 &              5 &   5 &   4 &   5 &   5 &   5 &   5 &    4 &    5 &    5 &    4 \\
DIS &         9 &         2 &         2 &         5 &             4 &   4 &   3 &   4 &   4 &   4 &   4 &    3 &    4 &    3 &    1 &              2 &   2 &   2 &   2 &   2 &   2 &   2 &    2 &    2 &    2 &    2 &              5 &   5 &   4 &   5 &   5 &   5 &   5 &    4 &    5 &    5 &    5 \\
ACU &         9 &         2 &         2 &         5 &             4 &   3 &   2 &   3 &   4 &   4 &   4 &    3 &    4 &    2 &    1 &              2 &   2 &   2 &   2 &   2 &   2 &   2 &    2 &    2 &    2 &    2 &              5 &   5 &   4 &   5 &   5 &   5 &   5 &    4 &    5 &    4 &    4 \\
WBM &         9 &         1 &         2 &         5 &             4 &   3 &   2 &   5 &   5 &   4 &   4 &    3 &    4 &    3 &    1 &              2 &   2 &   2 &   2 &   2 &   2 &   2 &    2 &    2 &    2 &    2 &              5 &   5 &   4 &   5 &   5 &   5 &   5 &    4 &    5 &    5 &    4 \\
FTS &         9 &         2 &         2 &         5 &             4 &   4 &   2 &   5 &   5 &   5 &   5 &    3 &    4 &    2 &    1 &              2 &   2 &   2 &   2 &   2 &   2 &   2 &    2 &    2 &    2 &    2 &              5 &   5 &   4 &   5 &   5 &   5 &   5 &    4 &    5 &    5 &    4 \\
\bottomrule
\end{tabular}
\label{tab:num_samples_linear}
\end{table}

\begin{table}
    \centering
    \tiny
    \caption{Number of samples run for all modes of the self-attention architecture. Additional samples will be run before any camera ready publication.}
\begin{tabular}{lrrrrrrrrrrrrrrrrrrrrrrrrrrrrrrrrrrrrr}
\toprule
{} &        MT &        ST &       FTD &       FTF & \multicolumn{11}{l}{ST Few-Shot \%} & \multicolumn{11}{l}{FTD Few-Shot \%} & \multicolumn{11}{l}{FTF Few-Shot \%} \\
{} & \multicolumn{4}{l}{Full-data} &           0.1 & 0.3 & 0.6 & 1.0 & 1.8 & 3.2 & 5.6 & 10.0 & 17.8 & 31.6 & 56.2 &            0.1 & 0.3 & 0.6 & 1.0 & 1.8 & 3.2 & 5.6 & 10.0 & 17.8 & 31.6 & 56.2 &            0.1 & 0.3 & 0.6 & 1.0 & 1.8 & 3.2 & 5.6 & 10.0 & 17.8 & 31.6 & 56.2 \\
\midrule
MOR &         7 &         2 &         1 &         3 &             4 &   3 &   3 &   3 &   3 &   3 &   3 &    3 &    3 &    2 &    3 &              1 &   1 &   1 &   1 &   1 &   1 &   1 &    1 &    1 &    1 &    1 &              3 &   3 &   3 &   3 &   2 &   2 &   1 &    1 &    1 &    1 &    1 \\
CMO &         7 &         2 &         1 &         2 &             3 &   4 &   4 &   4 &   4 &   4 &   4 &    3 &    3 &    3 &    2 &              1 &   1 &   1 &   1 &   1 &   1 &   1 &    1 &    1 &    1 &    1 &              2 &   2 &   2 &   2 &   2 &   2 &   2 &    1 &    1 &    1 &    1 \\
DNR &         7 &         2 &         1 &         3 &             4 &   3 &   3 &   3 &   3 &   3 &   3 &    3 &    2 &    3 &    2 &              1 &   1 &   1 &   1 &   1 &   1 &   1 &    1 &    1 &    1 &    1 &              3 &   3 &   3 &   2 &   2 &   1 &   1 &    1 &    1 &    1 &    1 \\
ICD &         7 &         2 &         1 &         3 &             4 &   3 &   3 &   3 &   3 &   3 &   3 &    3 &    4 &    4 &    2 &              1 &   1 &   1 &   1 &   1 &   1 &   1 &    1 &    1 &    1 &    1 &              3 &   3 &   3 &   3 &   2 &   1 &   1 &    1 &    1 &    1 &    1 \\
LOS &         7 &         2 &         1 &         3 &             3 &   4 &   3 &   3 &   3 &   3 &   3 &    3 &    2 &    2 &    2 &              1 &   1 &   1 &   1 &   1 &   1 &   1 &    1 &    1 &    1 &    1 &              3 &   3 &   3 &   3 &   3 &   2 &   2 &    1 &    1 &    1 &    1 \\
REA &         7 &         2 &         1 &         3 &             3 &   5 &   4 &   4 &   4 &   4 &   4 &    4 &    4 &    2 &    2 &              1 &   1 &   1 &   1 &   1 &   1 &   1 &    1 &    1 &    1 &    1 &              3 &   3 &   3 &   3 &   3 &   2 &   2 &    1 &    1 &    1 &    1 \\
DIS &         7 &         2 &         1 &         3 &             3 &   4 &   4 &   3 &   3 &   3 &   3 &    3 &    4 &    3 &    2 &              1 &   1 &   1 &   1 &   1 &   1 &   1 &    1 &    1 &    1 &    1 &              3 &   3 &   3 &   3 &   2 &   2 &   2 &    1 &    1 &    1 &    1 \\
ACU &         7 &         2 &         1 &         3 &             4 &   4 &   4 &   4 &   4 &   4 &   4 &    4 &    5 &    2 &    2 &              1 &   1 &   1 &   1 &   1 &   1 &   1 &    1 &    1 &    1 &    1 &              3 &   3 &   3 &   3 &   3 &   2 &   1 &    1 &    1 &    1 &    1 \\
WBM &         7 &         2 &         1 &         2 &             4 &   3 &   4 &   3 &   3 &   3 &   3 &    3 &    5 &    3 &    3 &              1 &   1 &   1 &   1 &   1 &   1 &   1 &    1 &    1 &    1 &    1 &              2 &   2 &   2 &   2 &   2 &   1 &   1 &    1 &    1 &    1 &    1 \\
FTS &         7 &         1 &         1 &         3 &             3 &   3 &   3 &   3 &   3 &   3 &   3 &    3 &    3 &    3 &    3 &              1 &   1 &   1 &   1 &   1 &   1 &   1 &    1 &    1 &    1 &    1 &              3 &   3 &   3 &   3 &   2 &   1 &   1 &    1 &    1 &    1 &    1 \\
\bottomrule
\end{tabular}
    \label{tab:num_samples_self_attention}
\end{table}

\end{document}